%% file: main.tex
\documentclass{article}
\usepackage{iclr2027_conference,times}
\usepackage[T1]{fontenc}
\usepackage{amsmath,amsfonts,bm}
\usepackage{graphicx}
\usepackage{float}
\usepackage{microtype}
\usepackage{url}
\usepackage{xspace}
\usepackage{hyperref}
\usepackage{enumitem}
\hypersetup{
  hidelinks,
  pdftitle={GLoD-LVSM: Learning Geometry-Grounded Continuous-LoD Scene Representations for Large View Synthesis Models},
  pdfauthor={Weikai Lin, Nicholas Milef, Carl S Marshall, Grant Kaijuin Yang, Todd Keeler, Yuhao Zhu}
}
\input{macro}
\iclrfinalcopy
\title{\proj{}: Learning Geometry-Grounded Continuous-LoD Scene Representations for Large View Synthesis Models}
\author{%
\parbox[t]{\dimexpr\textwidth-2\tabcolsep\relax}{\centering\normalfont\fontsize{12}{14}\selectfont
\makebox[\linewidth][c]{%
\textbf{\href{https://openreview.net/profile?id=~Weikai_Lin1}{Weikai Lin}}\textsuperscript{1,2}\hspace{1.25em}
\textbf{\href{https://openreview.net/profile?id=~Nicholas_Milef1}{Nicholas Milef}}\textsuperscript{2}\hspace{1.25em}
\textbf{\href{https://openreview.net/profile?id=~Carl_S_Marshall1}{Carl S Marshall}}\textsuperscript{2}\hspace{1.25em}
\textbf{\href{https://openreview.net/profile?id=~Grant_Kaijuin_Yang1}{Grant Kaijuin Yang}}\textsuperscript{2}}\\[4pt]
\textbf{\href{https://openreview.net/profile?id=~Todd_Keeler1}{Todd Keeler}}\textsuperscript{2}\hspace{6em}
\textbf{\href{https://openreview.net/profile?id=~Yuhao_Zhu1}{Yuhao Zhu}}\textsuperscript{1}\\[6pt]
\textsuperscript{1}University of Rochester\hspace{3em}\textsuperscript{2}Reality Labs, Meta
}
}
\begin{document}
\maketitle
\vspace{-12pt}
\lhead{Preprint}
\input{figures/teaser_layout}
\begin{abstract}
\input{sections/00_abstract}
\end{abstract}
\input{sections/01_introduction}
\input{sections/02_related_work}
\input{sections/03_method}
\input{sections/04_experiments}
\input{sections/05_conclusion}
\bibliography{references}
\bibliographystyle{iclr2027_conference}
\clearpage
\appendix
\setcounter{figure}{0}
\setcounter{equation}{0}
\renewcommand{\thefigure}{S\arabic{figure}}
\renewcommand{\theequation}{S\arabic{equation}}
\renewcommand{\theHfigure}{supp.\arabic{figure}}
\renewcommand{\theHequation}{supp.\arabic{equation}}
\input{sections/appendix}

\end{document}

%% file: macro.tex
\newcommand{\Sect}[1]{Sec.~\ref{#1}}
\newcommand{\Fig}[1]{Fig.~\ref{#1}}
\newcommand{\Eqn}[1]{Eqn.~\ref{#1}}
\newcommand{\proj}{\textsc{GLoD-LVSM}\xspace}
\newcommand{\mode}[1]{\underline{\textsc{#1}}\xspace}
\newcommand{\cp}{\mathcal{P}}
\newcommand{\cL}{\mathcal{L}}
\newcommand{\bi}{\mathbf{I}}
\def\va{{\bm{a}}}
\def\vz{{\bm{z}}}
\def\mZ{{\bm{Z}}}

%% file: figures/teaser_layout.tex
\begin{figure}[H]
\vspace{-10pt}
    \centering
    \includegraphics[width=0.95\linewidth]{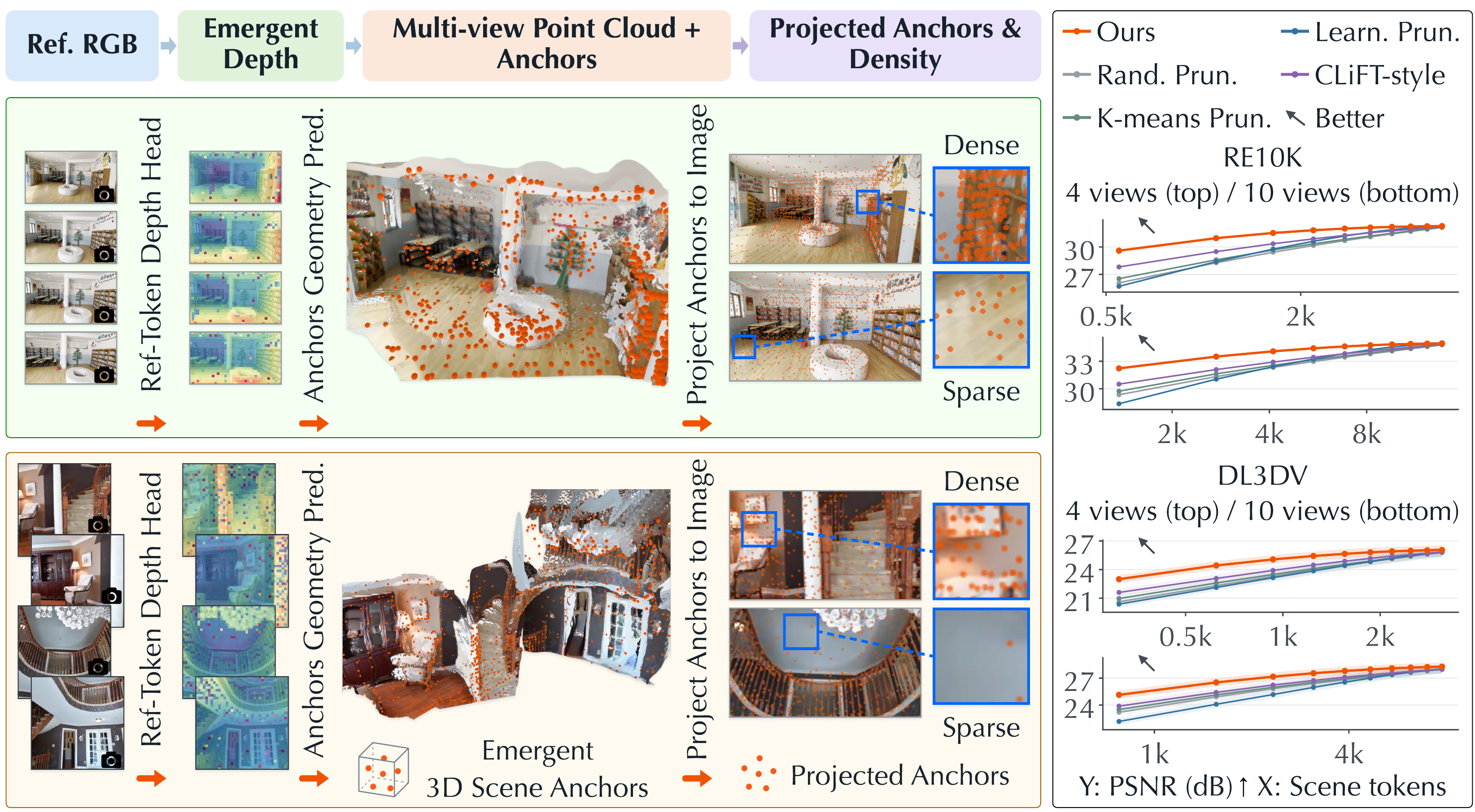}
    \caption{\proj learns a compact, geometry-grounded scene representation (anchors) for novel view synthesis. Left: View-synthesis supervision and compression alone yield emergent depth and self-organized, surface-aligned scene anchors whose density adapts to local texture complexity.
    Right: \proj achieves the best quality--scene-token trade-off across reference-view counts.}
    \label{fig:teaser}
    \vspace{-5pt}
\end{figure}

%% file: sections/00_abstract.tex
Large view synthesis models (LVSMs) reconstruct scenes from sparse views and render novel views quickly, but they incur unnecessary storage and rendering costs due to token redundancy.
We present \proj, a compact, geometry-grounded scene representation for LVSMs with continuous levels of detail.
First, instead of learning a fixed set of scene anchor initializations, \proj learns a continuous field that initializes a flexible number of anchors to match the desired budget.
A condenser then aggregates scene information into these anchors, with both the field and condenser trained end-to-end.
Second, the condenser infers reference-token and anchor geometry to guide compression through geometry-grounded attention, improving rendering quality and enabling frustum culling during decoding.
Compared with existing LVSM scene-compression baselines, \proj achieves the best trade-off between rendering quality and scene token count.
Compared with the uncompressed base model, \proj reduces the scene-token count by {33.4\%} and accelerates rendering by {$1.23\times$} with no quality loss.
Under a relaxed quality-loss budget, the reduction increases to {85.5\%}, with {$3.00\times$} rendering speedup.
Further analysis reveals that view-synthesis supervision yields emergent scene depth and self-organized anchors that align with surfaces and adapt their density to local texture complexity.
Beyond rendering, we further show that our scene representation also supports scene-semantic understanding,
highlighting its potential for broader downstream tasks.

%% file: sections/01_introduction.tex
\section{Introduction}
\label{sec:introduction}

Large view synthesis models (LVSMs) learn a reconstruction prior for novel view synthesis (NVS)~\citep{jin2025lvsm,kim2026svsm,szymanowicz2026lagernvs}, reducing scene reconstruction to a single feed-forward pass and enabling high-quality rendering from sparse reference views.
Recent high-quality LVSMs commonly adopt a highway representation, where each reference-image patch is directly mapped to a scene token~\citep{szymanowicz2026lagernvs,kim2026svsm}.
While this design preserves detailed appearance information, the resulting scene-token count grows with the number of reference views, increasing both storage and decoding costs. 
This tight coupling between image and scene-representation sizes becomes a bottleneck with more reference views.

In addition to scaling poorly with reference view count, highway representations allocate scene capacity inefficiently in two ways. 
First, overlapping 3D regions observed from different views are stored as separate tokens rather than consolidated. 
Second, every image patch receives the same one-token capacity regardless of texture complexity. 
An ideal LVSM representation should therefore (1) support flexible scene-token count independent of the input-view count, (2) consolidate multi-view observations into a unified, \textit{scene-centric} 3D representation, and (3) allocate capacity adaptively to scene complexity.
Thus, we propose \proj, which provides two key features: (1) a geometry-grounded scene representation that supports (2) continuous level-of-detail.
We refer to scene tokens in \proj as \textit{anchors} to emphasize their scene-centric 3D nature.

Different storage and rendering budgets require different scene-token numbers, which correspond to different levels of scene detail.
To support arbitrary scene-token counts and thus continuous levels of detail (CLoD), we first design an anchor initialization scheme that can generate a flexible number of tokens.
Specifically, we introduce a CLoD scene-anchor initialization that parameterizes anchor initialization as a continuous field (\Sect{sec:continuous-anchors}).
Sampling this field at different densities generates different numbers of anchors.
This decouples the scene-token count from the reference-view count and allows our model to operate under different storage and rendering budgets.

A compact scene representation should consolidate multi-view observations and allocate capacity where it is most useful.
To achieve this, we (1) use unified, scene-centric 3D anchors to consolidate redundant multi-view tokens, and (2) learn the spatial allocation of these anchors end-to-end.
Specifically, we use an end-to-end learned, geometry-grounded condenser that aggregates reference-view tokens into 3D anchors (\Sect{sec:condenser}).
The condenser predicts geometry for both views and anchors and uses their geometric relationships to guide cross-attention.
The resulting anchors have 3D positions, which further enable target-view frustum culling to reduce decoding cost (\Sect{sec:culling}).
Both the geometry and scene-complexity-aware anchor allocation emerge from the NVS objective (\Sect{sec:qualitative-analysis}).

We evaluate \proj across datasets, reference-view counts and scene-token budgets.
Quantitatively, \proj consistently achieves the best quality--token-count trade-off among the  baselines.
Qualitatively, the learned anchors emerge near scene surfaces and allocate higher density to regions with richer texture, suggesting that NVS supervision and compression alone can induce a geometry- and complexity-aware efficient scene representation.
Beyond rendering, we further evaluate the learned representation on scene-semantic understanding, demonstrating its potential as a 3D scene representation for broader downstream tasks.
Our contributions are threefold:
\begin{itemize}[leftmargin=2em]
\item \textbf{Continuous-LoD scene representation.}
We introduce a scene-conditioned continuous query field that enables flexible scene anchor budgets independent of the reference-view count.
\item \textbf{Geometry-grounded scene condensation.}
We learn scene-centric 3D anchors end-to-end, enabling efficient multi-view consolidation and scene-adaptive capacity allocation.
\item \textbf{Efficient and generalizable scene anchors.}
\proj achieves the best quality--scene-token trade-off for rendering and supports scene semantic understanding.
\end{itemize}

%% file: sections/02_related_work.tex
\section{Related Work}
\label{sec:related-work}

\subsection{Novel View Synthesis}
Novel view synthesis (NVS) is a widely studied problem and can be broadly categorized into per-scene optimization methods and feed-forward generalizable methods.
NeRF~\citep{mildenhall2020nerf} and 3D Gaussian Splatting (3DGS)~\citep{kerbl20233dgs} achieve photorealistic novel-view synthesis by optimizing a radiance field or a set of Gaussian primitives for each scene.
However, they typically require time-consuming per-scene optimization and dense image capture to achieve high rendering quality.
In contrast, generalizable NVS methods~\citep{sajjadi2022srt,chen2024mvsplat,xu2025depthsplat}, including recent Transformer-based approaches such as LVSMs~\citep{jin2025lvsm,kim2026svsm,sun2026dvsm,szymanowicz2026lagernvs}, learn reconstruction priors from large datasets and can reconstruct unseen scenes from sparse input views in a single feed-forward pass, avoiding costly dense image capture and per-scene optimization.
Our work builds on Transformer-based generalizable NVS and focuses on learning a compact, geometry-grounded scene representation that reduces redundancy among scene tokens, thereby lowering storage and rendering costs.

\subsection{Compact Scene Representation}

Compact scene representations have been widely studied for efficient NVS.
For example, 3DGS methods prune or compress Gaussian primitives to reduce representation size while preserving visual quality~\citep{fan2024lightgaussian,fang2024mini,lin2025metasapiens}.
Some methods further support multiple or continuous levels of detail (CLoD), allowing representation complexity to adapt to runtime rendering budgets~\citep{milef2025learning,cheng2026clodgs}.
Compact scene-token representations have also been explored for LVSMs.
LVSM-ED~\citep{jin2025lvsm} uses a fixed latent bottleneck and cannot support flexible compression rates.
CLiFT~\citep{wang2025clift} uses K-means to select representative reference-view tokens, relying on handcrafted clustering that may be suboptimal.
Our work also targets compact scene representations for LVSMs. Unlike prior methods, we support CLoD and learn scene anchors end-to-end, enabling flexible compression with less handcrafted design.

%% file: sections/03_method.tex
\section{\proj: Geometry-Grounded CLoD Scene Representation}
\label{sec:method}

\begin{figure}[t]
    \centering
    \includegraphics[width=\linewidth]{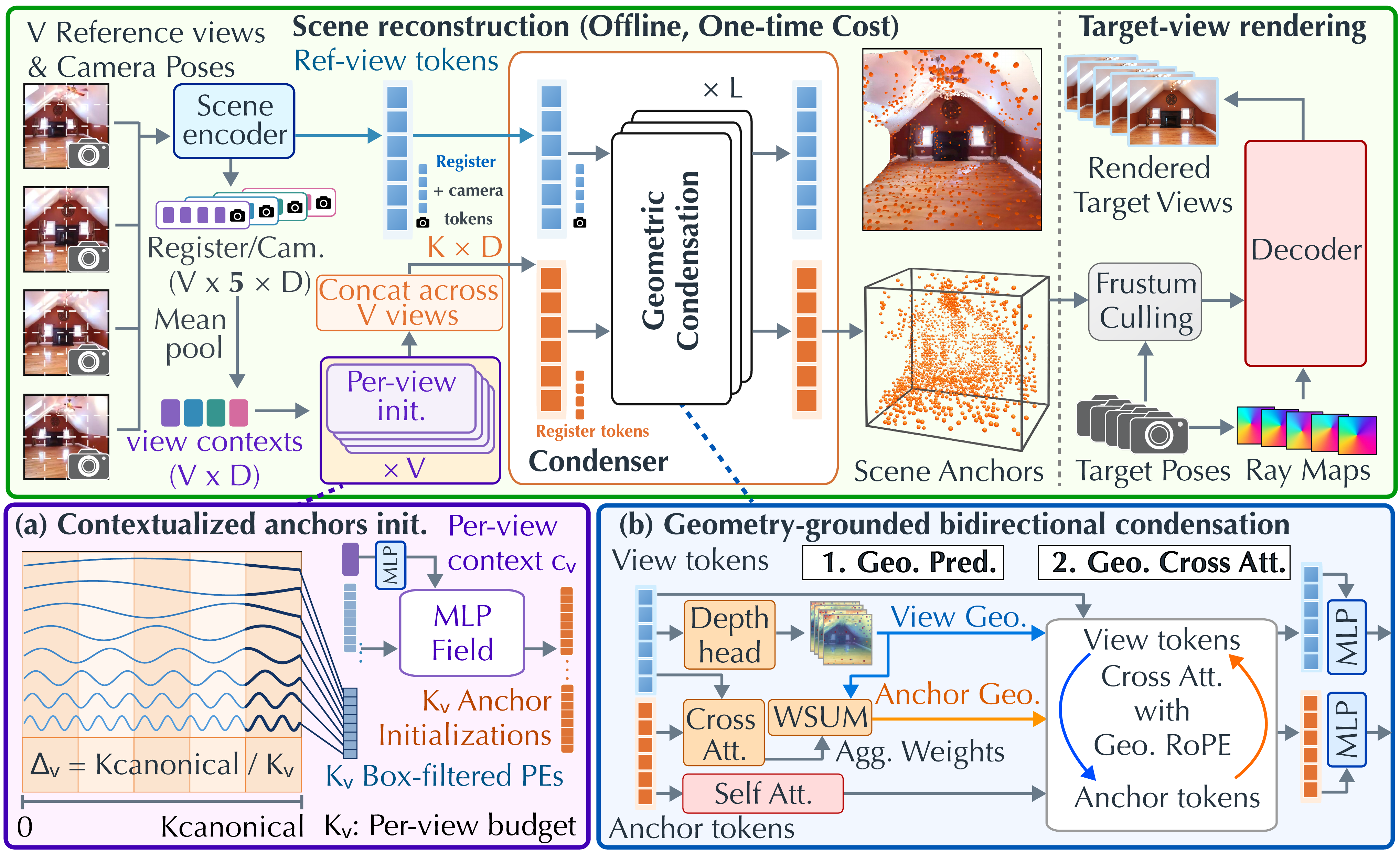}
    \caption{Overview of \proj.
    A highway encoder extracts reference-view tokens and per-view context.
    (a) A scene-conditioned field uses box-filtered positional encodings to initialize the chosen number of anchors.
    (b) The condenser predicts geometry for both reference-view tokens and anchors and uses geometry-grounded bidirectional attention to condense reference-view tokens into anchors.
    The same scene anchors are reused to render different target views.}
    \label{fig:method-overview}
    \vspace{-5pt}
\end{figure}

\Fig{fig:method-overview} illustrates \proj.
Given $V$ reference images and their camera parameters, \proj constructs a compact scene representation with a chosen budget of $K$ anchors.
A highway scene encoder first extracts reference-view tokens (\Sect{sec:scene-encoder}).
Our anchor initializer then generates $K$ anchors from a  continuous field (\Sect{sec:continuous-anchors}).
These anchors are passed to our condenser, which produces scene-centric 3D anchors (\Sect{sec:condenser}).
The anchors will be used to render the RGB images given the target views (\Sect{sec:anchor-rendering}).
We train \proj end-to-end using the NVS loss (\Sect{sec:training}).

\subsection{Highway Scene Encoder}
\label{sec:scene-encoder}

We use the encoder from LagerNVS~\citep{szymanowicz2026lagernvs}.
The inputs are $\mathcal{X}=\{(\bi_v,\cp_v)\}_{v=1}^{V}$, where $\bi_v$ is the $v$-th reference image and $\cp_v$ is its camera parameters.
The encoder patchifies each reference image, embeds each patch as a token, embeds $\cp_v$ into a camera token $\mathbf{t}^{\mathrm{cam}}_v$, and appends four register tokens $\mathbf{t}^{\mathrm{reg}}_{1:4}$ to each view~\citep{darcet2024registers}.
For each view, input token sequence
$\mathbf{T}_v^{\mathrm{in}}=
[\mathbf{t}^{\mathrm{cam}}_v,
\mathbf{t}^{\mathrm{reg}}_{1:4},
\operatorname{PatchEmbed}(\bi_v)]$.
The resulting tokens are jointly encoded across all $V$ views:
\begin{equation}
    \mathbf{H}=\operatorname{Encoder}(\{\mathbf{T}_v^{\mathrm{in}}\}_{v=1}^{V}),
    \qquad
    \mZ=\mathbf{H}_{\mathrm{patch}}\mathbf{W}\in\mathbb{R}^{N\times D}.
    \label{eq:scene-encoder}
\end{equation}
Here, $\mathbf{W}$ linearly maps the encoded patch tokens to a defined $D$-dimensional space, and $N$ is the total number of input patches across all views.
We refer to the output $\mZ$ as \emph{reference-view tokens} and retain the encoded register and camera tokens for anchor initialization (\Sect{sec:continuous-anchors}).
This one-patch-one-token highway design causes storage and decoding costs to grow with reference-view count.

\subsection{Contextualized Continuous Field for Anchor Initialization}
\label{sec:continuous-anchors}

Given reference-view tokens $\mZ$, our condenser compresses them into scene anchors.
The first challenge is anchor initialization.
Fixed learned vectors~\citep{jin2025lvsm} support only a predefined anchor count.
Clustering-based initialization~\citep{wang2025clift} supports variable counts, but relies on handcrafted clustering that may be suboptimal and remains image-centric rather than scene-centric.
We instead learn a scene-conditioned continuous initialization field end-to-end with minimal handcrafted design (\Fig{fig:method-overview}(a)).
Sampling the field at different densities yields different anchor budgets, enabling continuous levels of detail under varying storage and rendering constraints.

\paragraph{Continuous field parameterization.}
Given a total anchor budget $K$, we distribute $K$ evenly across the $V$ reference views, assigning $K_v$ anchors to view $v$.
For each view, our goal is to learn a $D$-dimensional continuous field that maps query indices $i\in\{0,\ldots,K_v-1\}$ to $K_v$ vectors in the $D$-dimensional anchor space, with $D$ set to the highway model's scene-token dimension.
We parameterize this continuous field as an MLP conditioned jointly on the anchor indices and the current scene/view context. 
Specifically, each anchor index $i$ is mapped to a continuous coordinate $u_{v,i}$ and encoded with our anti-aliased positional encoding $\bar{\gamma}(u_{v,i})$ (\Eqn{eq:anchor-intervals}). 
To condition the field on the current input, we mean-pool the encoded registers and camera tokens (\Sect{sec:scene-encoder}) of view $v$ and map them through $\operatorname{MLP}_{\mathrm{context}}$ to obtain a view- and scene-specific context vector $\mathbf{c}_v$. 
$\bar{\gamma}(u_{v,i})$ and $\mathbf{c}_v$ are then concatenated and passed to $\operatorname{MLP}_{\mathrm{field}}$ to produce the $D$-dimensional anchor initialization $\va_{v,i}$:
\begin{equation}
    \va_{v,i}
    =
    \operatorname{MLP}_{\mathrm{field}}
    \left(
    \left[
    \bar{\gamma}(u_{v,i}),
    \mathbf{c}_v
    \right]
    \right),
    \qquad
    \mathbf{c}_v
    =
    \operatorname{MLP}_{\mathrm{context}}
    \left(
    \operatorname{MeanPool}([\mathbf{H}^{\mathrm{reg}}_{v,1:4},\mathbf{H}^{\mathrm{cam}}_v])
    \right).
    \label{eq:contextual-query}
\end{equation}
Finally, we concatenate the initializations from all views to obtain
$K=\sum_v K_v$ anchor initializations and feed them to the condenser.
The field is optimized end-to-end through the NVS objective.

\paragraph{Anti-Aliased Positional Encoding Across Anchor Budgets.}

To sample the same continuous field at different anchor budgets, we map each discrete query index $i$ to a continuous coordinate $u_{v,i}$ in a fixed canonical domain $[0,K_{\mathrm{canonical}}]$, where $K_{\mathrm{canonical}}=2000$.
For each view with budget $K_v$, we divide the domain into $K_v$ intervals.
Changing $K_v$ therefore changes the sampling density while preserving coverage of the domain.
To help the field capture high-frequency variation, we follow NeRF-style positional encoding (PE)~\citep{mildenhall2020nerf} and encode each coordinate as $\gamma(u)=\frac{1}{\sqrt{F}}[\cos(2\pi\mathbf{f}u),\sin(2\pi\mathbf{f}u)]\in\mathbb{R}^{2F}$.
Here, $\mathbf{f}=(f_0,\ldots,f_{F-1})$ is logarithmically spaced between
$f_{\min}=1/20000$ and $f_{\max}=0.5$.
We set $F=D/2$, where $D$ is the dimension of anchor space.
At small budgets, sparse sampling can alias the high-frequency components of PE.
We therefore integrate PE over the sampling interval, analogous to box filtering~\citep{barron2021mipnerf}:
\begin{equation}
    \bar{\gamma}_{v,i}=\frac{1}{\Delta_v}\int_{I_{v,i}}\gamma(u)\,du,\quad
    I_{v,i}=[u_{v,i},u_{v,i}+\Delta_v],\quad
    u_{v,i}=i\Delta_v,\quad
    \Delta_v=\frac{K_{\mathrm{canonical}}}{K_v}.
    \label{eq:anchor-intervals}
\end{equation}
As the anchor budget decreases, the intervals become wider and more strongly attenuate high-frequency components.
The sine and cosine integrals have closed-form solutions, given in Supplementary~\ref{supp:expected-frequency-derivation}.
The resulting $\bar{\gamma}(u_{v,i})\in\mathbb{R}^{D}$ provides a $D$-dimensional encoding of each anchor index whose frequency content reflects the anchor budget, with lower budgets inducing stronger attenuation.

\subsection{Geometry-Grounded Bidirectional Condensation}
\label{sec:condenser}
As illustrated in \Fig{fig:method-overview}(b), our condenser maps $K$ initialized anchors and $N$ reference-view tokens $\mZ$ to $K$ compact scene anchors.
It uses $L=8$ bidirectional Transformer blocks~\citep{szymanowicz2026lagernvs}, where the two streams attend to each other.
Content-only attention can mix spatially unrelated information because it lacks 3D awareness.
We therefore predict geometry for both streams (\Sect{sec:view-token-geometry}) and use it to modulate the bidirectional cross-attention (\Sect{sec:geometry-attention}).
In the condenser, we retain the reference-view register and camera tokens and add four anchor registers. 
These tokens also participate in attention, but with no geometry modulation.

\subsubsection{Predicting View and Anchor Geometry}
\label{sec:view-token-geometry}
\paragraph{View Geometry.}
We predict geometry for each reference-view token directly from its feature.
Following RayRoPE~\citep{wu2026rayrope}, a linear depth head maps token $\vz_j$ to a log-depth center $\mu_j$ and an uncertainty scale $\sigma_j$.
Using the known camera parameters, we back-project the four corners of the corresponding image patch at depth
$d_j=e^{\mu_j}$ to obtain its spatial span in 3D space
$\mathbf{p}_{j,1:4}$.
We also use the unit center-ray direction $\mathbf{v}_j$ to encode its viewing direction.
Together, $\{\mathbf{p}_{j,1:4},\mathbf{v}_j,\sigma_j\}$ describes the spatial support and viewing direction of reference-view token $j$.
The geometry is recomputed from the updated reference-view features at every condenser layer.

\paragraph{Anchor Geometry.}
Rather than directly regressing unconstrained 3D coordinates for each anchor, we derive its geometry by aggregating the geometry of the reference-view tokens. 
This introduces a useful prior: it constrains anchors to the convex hull of the observed geometry rather than allowing unconstrained 3D positions.
Specifically, we use a lightweight single-head cross-attention  to compute aggregation weights $\beta_{ij}$ between anchor $i$ and reference-view token $j$:
\begin{equation}
\beta_{ij}
=
\operatorname{softmax}_j
\left(
\frac{\hat{\mathbf{q}}_i^\top\hat{\mathbf{k}}_j}{\sqrt{64}}
\right),
\qquad
\mathbf{p}_i^{A}
=
\sum_j \beta_{ij}\bar{\mathbf{p}}_j,
\qquad
\mathbf{v}_i^{A}
=
\sum_j \beta_{ij}\mathbf{v}_j,
\label{eq:anchor-geometry}
\end{equation}
where $\hat{\mathbf{q}}_i$ and $\hat{\mathbf{k}}_j$ are 64-dimensional query and key projections, and
$\bar{\mathbf{p}}_j=\tfrac{1}{4}\sum_{c=1}^{4}\mathbf{p}_{j,c}$
is the center of the back-projected view-token patch.
The same weights aggregate both position and viewing direction, giving each anchor a 3D position $\mathbf{p}_i^{A}$ and a viewing-direction descriptor $\mathbf{v}_i^{A}$.
$\beta_{ij}$ is normalized over the reference-view tokens.
Importantly, we use the anchor geometry predicted at the final condenser block as the anchor's final geometry for subsequent rendering.

\subsubsection{Geometry-Grounded Bidirectional Attention}
\label{sec:geometry-attention}

After geometry prediction, each condenser block first applies self-attention to the anchor stream, followed by bidirectional cross-attention between anchors and reference-view tokens and MLPs.
We inject the inferred geometry into this cross-attention using geometry-dependent RoPE.

We define the geometric descriptors for the two token streams as follows.
A reference-view token $j$ corresponds to an image patch and is represented by its four back-projected patch corners $\mathbf{p}_{j,1:4}$, viewing direction $\mathbf{v}_j$, and depth uncertainty $\sigma_j$.
An anchor $i$, in contrast, has a single 3D position $\mathbf{p}_i^A$ and viewing-direction descriptor $\mathbf{v}_i^A$.
To match the geometric layout of the reference-view tokens, we repeat $\mathbf{p}_i^A$ four times for each anchor.
Following RayRoPE~\citep{wu2026rayrope}, we represent a 3D point using $\pi(\mathbf{p})$, which first transforms the point into the reference camera frame to obtain $(X,Y,Z)$, then computes $\pi(\mathbf{p})=(X/r,Y/r,1/|Z|)$ with $r=\sqrt{X^2+Y^2+Z^2}$.
The four points therefore provide 12 coordinates.
Appending the 3D viewing-direction descriptor gives a 15-dimensional geometric descriptor for both reference-view tokens and anchors:
\begin{equation}
    \mathbf{g}^{R}_j
    =
    [\pi(\mathbf{p}_{j,1}),\ldots,\pi(\mathbf{p}_{j,4}),\mathbf{v}_j],
    \qquad
    \mathbf{g}^{A}_i
    =
    [\pi(\mathbf{p}^{A}_i),\ldots,\pi(\mathbf{p}^{A}_i),\mathbf{v}^{A}_i]
    \in\mathbb{R}^{15}.
    \label{eq:geometry-descriptors}
\end{equation}

For reference-view token geometry $\mathbf{g}^{R}_j$, we use RayRoPE's expected RoPE as encoding to account for the predicted depth uncertainty $\sigma_j$ (\Sect{sec:view-token-geometry}).
For anchor geometry $\mathbf{g}^{A}_i$
, we use standard RoPE encoding~\citep{su2024roformer}.
The encodings are then used to rotate the query, key, value, and output features in the bidirectional cross-attention, following the rotation mechanism in RayRoPE.

A key challenge is choosing the RoPE coordinate frame.
When an anchor interacts with tokens from a reference view, we express RoPE in that view's frame for both cross-attention directions.
This avoids dependence on the arbitrary world coordinate system, similar to conventional graphics pipelines that transform scene geometry into camera space before rasterization~\citep{akenine2018realtime}.

\subsection{Runtime Culling and Rendering}
\label{sec:anchor-rendering}
\label{sec:culling}

\paragraph{Culling.}
After scene reconstruction, the same set of anchors can be reused across different target views.
We use the anchor geometry from the final condenser layer as the geometry of each anchor.
Given a target camera $\cp_t$, we project each anchor position $\mathbf{p}_i^A$ into the target camera frame and obtain its normalized image coordinate $(u_i^t,v_i^t)$ and camera-space depth $z_i^t$.
We retain anchors that lie in front of the camera and inside a slightly padded image region, $\mathcal{K}_t=\{i:z_i^t>0,\;(u_i^t,v_i^t)\in[-\delta,1+\delta]^2\}$, with $\delta=0.1$.
The padding retains anchors close to the image boundary.
Only anchors in $\mathcal{K}_t$ are passed to the decoder, reducing the number of scene tokens processed for each target view.

\paragraph{Novel-View Rendering.}
We use the bidirectional decoder of LagerNVS~\citep{szymanowicz2026lagernvs}, modified with our geometry-grounded RoPE for anchor--target-token cross-attention (\Sect{sec:geometry-attention}).
The target camera is converted into a ray map and patch-embedded into target tokens, which predict their view geometry with a linear head (as for reference-view tokens) and interact with anchors through bidirectional cross-attention.
Anchors use the fixed geometry from the final condenser layer.
The target camera frame is used as the RoPE coordinate frame.
Finally, a linear projection maps target tokens to RGB patches, which form the rendered image $\hat{\bi}_t$.

\subsection{Training}
\label{sec:training}

We freeze the pretrained scene encoder and jointly optimize the continuous query field, condenser, and decoder.
The decoder is supervised by the LagerNVS rendering loss $\cL_{\mathrm{render}}$~\citep{szymanowicz2026lagernvs}.
We additionally regularize the predicted depth uncertainty $\sigma_j^\ell$ of reference-view token $j$ at condenser layer $\ell$ (\Sect{sec:view-token-geometry}) to encourage sharp, localized geometry:
\begin{equation}
    \cL
    =
    \cL_{\mathrm{render}}
    +\lambda_\sigma\cL_{\mathrm{decay}},
    \qquad
    \cL_{\mathrm{decay}}
    =
    \frac{1}{LN}
    \sum_{\ell=0}^{L-1}
    \sum_{j=1}^{N}
    |\sigma_j^\ell|,
    \qquad
    \lambda_\sigma=10^{-3}.
    \label{eq:training-loss}
\end{equation}
We first train without culling, then enable culling and perform culling-aware fine-tuning.
During training, we sample 1--10 reference views and uniformly sample the anchor budget from 256 to the corresponding highway-token count $N$.
No explicit depth or anchor-position supervision is used.

%% file: sections/04_experiments.tex
\section{Evaluation}
\label{sec:experiments}

\subsection{Experimental Setup}
\label{sec:experimental-setup}

\paragraph{Datasets.}
We evaluate on RealEstate10K~\citep{zhou2018stereomag} and DL3DV~\citep{ling2024dl3dv} at resolutions of $256\times256$ and $256\times448$, respectively.
We study varying-reference-view settings from $2$ to $10$ views.
Since the official protocols cover only 2 views for RealEstate10K and 2/4/6 views for DL3DV, we construct the higher-view settings using the method described in Supplementary~\ref{supp:evaluation-details}.

\paragraph{Training.}
We fine-tune the pretrained general LagerNVS checkpoint~\citep{szymanowicz2026lagernvs} on each dataset using its $1$--$10$-view training protocol, obtaining \mode{LagerNVS-FT}, which is our base model.
We initialize our decoder from \mode{LagerNVS-FT} and train the condenser and decoder, followed by culling-aware fine-tuning.
See Supplementary~\ref{supp:finetuning-details} for training details.

\paragraph{Baselines.}
We compare against two groups of baselines:
\begin{itemize}[leftmargin=1.5em]
    \item \textbf{System-level baselines.} We compare \proj against \mode{CLiFT}~\citep{wang2025clift}, \mode{LVSM}~\citep{jin2025lvsm}, \mode{MVSplat}~\citep{chen2024mvsplat}, and \mode{DepthSplat}~\citep{xu2025depthsplat}.
    \item \textbf{Condenser baselines.} 
    Alternative condensers using the same base model and training schedule:
    \begin{itemize}[leftmargin=1.5em]
        \item \mode{Random pruning}: randomly selects scene tokens to match the desired budget.
        \item \mode{K-means pruning}: sets the number of clusters to the desired token budget, clusters the reference-view token features, and retains only the token closest to each cluster centroid.
        \item \mode{Learned Pruning}: uses a trained token pruner to select 90\% of the token budget, then fills the remaining 10\% through feature-diversity-based sampling~\citep{takezoe2026learnpruner}.
        \item \mode{CLiFT-style}: uses a K-means-based Transformer condenser adapted from \mode{CLiFT}~\citep{wang2025clift}, with the number of blocks increased to eight to match our condenser.
    \end{itemize}
\end{itemize}

\paragraph{Metrics.}
We report novel-view PSNR, SSIM~\citep{wang2004ssim}, and LPIPS~\citep{zhang2018lpips} under different reference view counts and scene token budgets.
For all quality metrics, we report the mean and 95\% confidence intervals (CIs) from 2,000 scene-level resamples with replacement.

\begin{figure}[t]
\centering
\begin{minipage}[t]{0.49\linewidth}
\vspace{0pt}
\centering
\includegraphics[width=\linewidth]{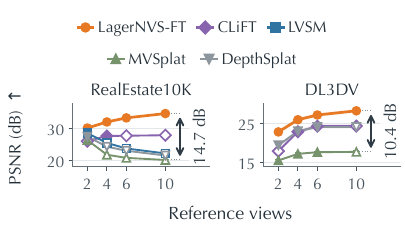}
\caption{PSNR across datasets and view counts. Shaded bands show 95\% CI.}
\label{fig:base-model}
\end{minipage}\hfill
\begin{minipage}[t]{0.49\linewidth}
\vspace{0pt}\centering
\includegraphics[width=\linewidth]{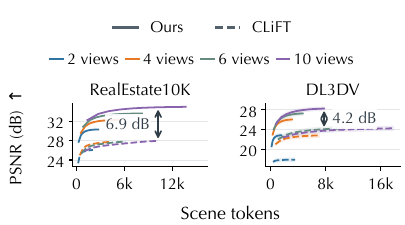}
\caption{PSNR vs. scene-token counts.
Shaded bands show 95\% CI.}
\label{fig:system-tradeoff}
\end{minipage}
\vspace{-5pt}
\end{figure}

\subsection{Comparing against other Systems}
\label{sec:base_model_comp}

\paragraph{Our base model is strong.}
To show that we build on a strong highway backbone, we first compare our base model, \mode{LagerNVS-FT}, against other systems.
\Fig{fig:base-model} shows the results, where filled markers denote view counts seen during training and hollow markers denote unseen counts.
\mode{LagerNVS-FT} consistently achieves substantially higher PSNR than all compared systems\footnote{LVSM is evaluated only on RealEstate10K because no official DL3DV checkpoint is available.}; SSIM and LPIPS show similar trends (Supplementary \Fig{fig:base-model-additional-metrics}).
It also improves as more reference views are provided.
\footnote{DL3DV results in \Fig{fig:base-model} are not directly comparable across reference-view counts because the official target set is defined as the complement of the reference views and therefore changes slightly with view count.}

\paragraph{\proj achieves the best quality--token trade-offs.}
We compare the quality--scene-token trade-off with \mode{CLiFT} because all other systems do not support flexible scene-token counts.
\Fig{fig:system-tradeoff} shows the results for different reference views.
\proj achieves a better PSNR--token-count trade-off than \mode{CLiFT}.
The advantage comes from both our strong base model and our better condenser design.
SSIM and LPIPS show similar trends (see Supplementary \Fig{fig:system-additional-metrics}).

\subsection{Comparing against Different Condenser Designs}
\label{sec:compression-results}
\begin{figure}[t]
\centering
\includegraphics[width=\linewidth]{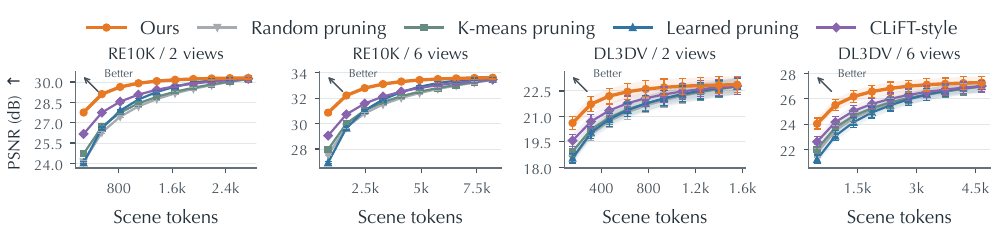}
\caption{Comparison of condenser quality--token trade-offs using the same base model.
Left: RealEstate10K with 2 and 6 reference views.
Right: DL3DV with 2 and 6 reference views.
The x-axis denotes the pre-culling scene-token count.
Shaded bands indicate 95\% confidence intervals.
}
\label{fig:compression}
\vspace{-5pt}
\end{figure}

\begin{figure}[t]
\centering
\includegraphics[width=\linewidth]{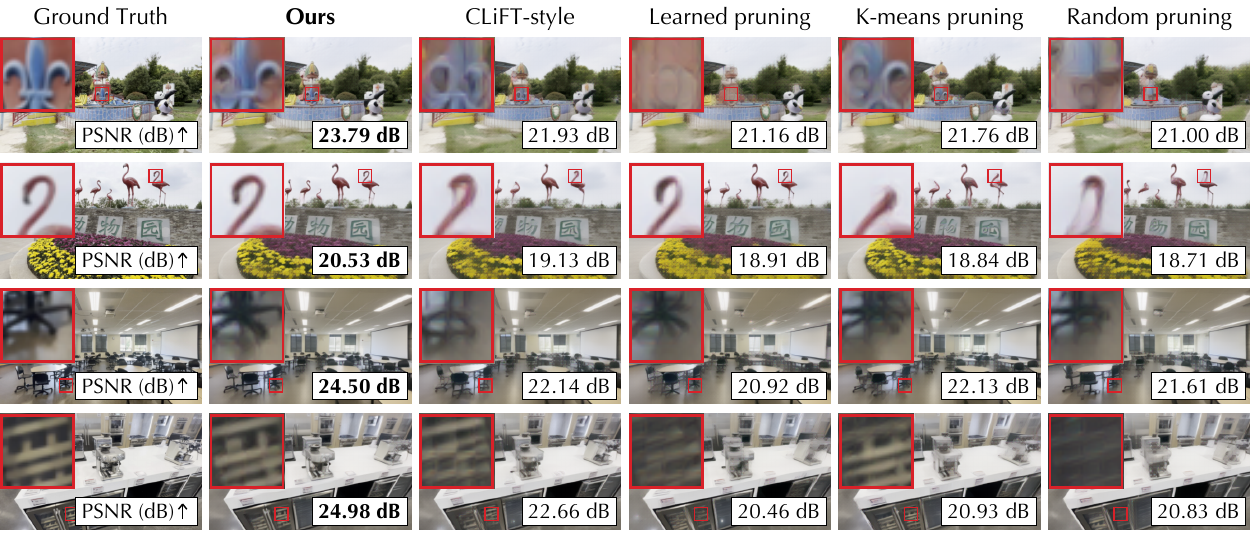}
\caption{Visualization of novel-view predictions from different condensers using two reference views at the same budget of 311 scene tokens (80\% reduction), shown on four DL3DV scenes.}
\label{fig:compression-visual-main}
\vspace{-5pt}
\end{figure}

To isolate the effect of the strong base model, we conduct controlled experiments using the same base model and training schedule across different condenser designs.
Because the baseline condensers do not produce geometry-grounded tokens, applying frustum-culling is not straightforward.
For a fair comparison, we disable culling for all methods and report the compressed scene-token count.
\Fig{fig:compression} compares quality--token trade-offs with 2 and 6 reference views on both datasets; 4- and 10-view results are shown in \Fig{fig:teaser}.
Our condenser achieves a better PSNR--token-count trade-off across datasets and reference-view counts.
SSIM and LPIPS show similar trends (see Supplementary Figs.~\ref{fig:condenser-additional-metrics}--\ref{fig:condenser-additional-10views}).
We also visualize novel-view predictions from all condensers in \Fig{fig:compression-visual-main}, which visually confirms our method's quality advantage.
More visualizations across scenes, reference-view counts, and compression rates are provided in Supplementary Figs.~\ref{fig:compression-visual-07}--\ref{fig:compression-visual-06}.

\subsection{Token Reduction \& Rendering Speedup over the Base Model}
\label{sec:quality-budget-savings}
\begin{figure}[t]
\centering
\begin{minipage}[t]{0.23\linewidth}
\vspace{0pt}\centering
\includegraphics[width=\linewidth]{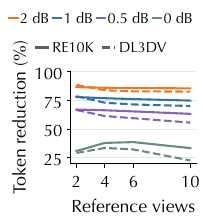}
\raggedright
\caption{Scene-token reductions at different PSNR  loss budgets.}
\label{fig:quality-budget-token-savings}
\end{minipage}\hfill
\begin{minipage}[t]{0.23\linewidth}
\vspace{0pt}\centering
\includegraphics[width=\linewidth]{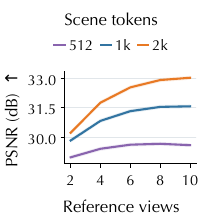}
\raggedright
\caption{Scaling quality with reference views on RealEstate10K.}
\label{fig:fixed-budget-psnr}
\end{minipage}\hfill
\begin{minipage}[t]{0.23\linewidth}
\vspace{0pt}\centering
\includegraphics[width=\linewidth]{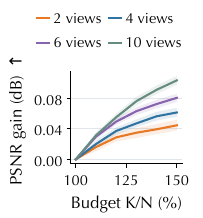}
\raggedright
\caption{Scaling beyond the base-model token budget on DL3DV.}
\label{fig:beyond-input-psnr}
\end{minipage}\hfill
\begin{minipage}[t]{0.23\linewidth}
\vspace{0pt}\centering
\includegraphics[width=\linewidth]{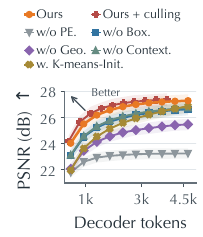}
\raggedright
\caption{DL3DV ablation studies using 6 reference views.}
\label{fig:ablation-tradeoff}
\end{minipage}
\vspace{-5pt}
\end{figure}

We now evaluate the token reduction and rendering speedup against the base model.
Efficiency gains must be considered in the context of rendering quality.
Using \mode{LagerNVS-FT} as the quality reference, we measure scene-token reduction under PSNR losses of $X\in\{2,1,0.5,0\}$\,dB.
\Fig{fig:quality-budget-token-savings} shows the results. 
On 10-view RealEstate10K, \proj reduces the scene-token count by 85.5\% at a 2 dB quality loss. Even with no quality loss (0 dB), it still reduces the token count by 33.4\%, which we attribute to the more effective geometry-grounded representation. The corresponding rendering speedups are $3.00\times$ and $1.23\times$ over \mode{LagerNVS-FT}, respectively. With culling, the estimated speedups further increase to $3.20\times$ and $1.55\times$.
More speed results and evaluation details are provided in Supplementary \Fig{fig:rendering-speed} and Supplementary~\ref{supp:speed-protocol}.

\subsection{Scaling Quality with Views and Beyond the Highway Token Count}
\label{sec:fixed-budget-scaling}
We now highlight two properties of \proj.
First, unlike highway designs whose scene tokens grow with reference-view count, \proj can exploit more views  at a \textit{fixed scene-token budget}.
\Fig{fig:fixed-budget-psnr} shows that, at a fixed token budget, \proj generally improves with more views. 
For example, on RealEstate10K with a 2048-token budget, 10 views outperform 2 views by 2.8 dB.
In contrast, a 10-view highway model requires over 13K tokens.
At small capacity, such as 512 tokens, gains saturate/degrade beyond 6 views.
See Supplementary Fig.~\ref{fig:fixed-budget-both-datasets} for more results.

Second, because \proj learns a continuous anchor-initialization field (\Sect{sec:continuous-anchors}), it can sample more anchors ($K$) than the original highway representation tokens ($N$), which baseline condensers cannot support.
\Fig{fig:beyond-input-psnr} reports the PSNR gain relative to \proj at the uncompressed budget.
Although \proj exceeds the base-model quality at the same token budget, its quality continues to improve for $K > N$.
See Supplementary Fig.~\ref{fig:beyond-input-additional-metrics} for more results.

\subsection{Emergent Geometry and Anchor Allocation}
\label{sec:qualitative-analysis}
To understand \proj, we visualize its representation using four reference views with 2048 scene anchors (\Fig{fig:teaser}).
Without explicit geometry supervision, consistent scene depth emerges across reference views, while the learned anchors concentrate near scene surfaces.
Projecting the anchors back onto the reference images further reveals scene-adaptive allocation: anchors are denser in texture-rich regions and sparser in textureless regions.
These results show that the NVS loss can induce an efficient geometry- and texture-aware scene representation.
More visualizations are in Supplementary Figs.~\ref{fig:qualitative-re10k-1}--\ref{fig:qualitative-dl3dv-3}.
We process raw depth before point-cloud visualization; see Supplementary~\ref{supp:depth-visualization}.

\subsection{Ablation Studies}
\label{sec:culling-evaluation}
\label{sec:geometry-analysis}
\Fig{fig:ablation-tradeoff} compares quality--decoder-token trade-offs for our ablations on 6-view DL3DV.
\mode{Ours} denotes \proj without culling.
\mode{Ours + Culling} enables our frustum-culling (\Sect{sec:culling}).
\mode{w/o Geo.} removes geometry-dependent RoPE (\Sect{sec:geometry-attention}) from both the condenser and decoder.
\mode{w/o Box.} removes anti-aliasing from the positional encoding (\Eqn{eq:anchor-intervals}).
\mode{w/o PE.} removes positional encoding and instead feeds normalized anchor indices to the field MLP.
\mode{w/o Context.}\footnote{Removing the context vector makes the per-view initialization identical across views. We therefore replace the per-view fields with a single global initialization field and set $K_{\mathrm{canonical}}=20{,}000$.} removes scene- and view-specific conditioning from the anchor initialization field (\Eqn{eq:contextual-query}).
\mode{w. K-means Init.} replaces our end-to-end learned anchor initialization field (\Sect{sec:continuous-anchors}) with K-means centroids of the reference-view tokens~\citep{wang2025clift}.
The results show that each component contributes to the quality--token trade-off of \proj.
Results with other reference-view counts show similar trends and are provided in Supplementary Figs.~\ref{fig:ablation-supp-2-4views}--\ref{fig:ablation-supp-6-10views}.

\subsection{\proj Supports Scene Semantic Understanding}
\label{sec:beyond-rendering}

We further examine whether the learned anchors capture scene semantics beyond rendering.
We use VLM-generated annotations for RealEstate10K~\citep{he2024cameractrl,li2023lavis} to categorize scenes into eight semantic classes.
For each scene, \proj compresses 10 reference views into 1,024 anchors, which are mean-pooled into a single 768-dimensional \textit{scene vector}.
We train only a linear probe on this vector for scene classification and achieve 80.40\%/91.73\% top-1/top-2 accuracy.
This result shows that learned anchors already contain scene semantics.
See Supplementary~\ref{supp:semantic-probe} for evaluation protocol and
Supplementary \Fig{fig:semantic-probe} for the confusion matrix.

%% file: sections/05_conclusion.tex
\section{Conclusion}
\label{sec:conclusion}

We propose \proj, a compact, geometry-grounded LVSM representation with continuous levels of detail.
\proj achieves a strong quality--token trade-off through two key designs: (1) an end-to-end learned continuous field with minimal handcrafted design, and (2) a condenser that consolidates multi-view observations into scene-centric, geometry-grounded anchors.
Trained with the NVS objective, \proj learns geometry- and texture-aware anchor allocation.
We further show its utility on scene semantic understanding, highlighting the potential beyond rendering.

%% file: sections/appendix.tex
\begin{center}
    {\Large\bfseries \proj: Supplementary Material}
\end{center}

\section{Implementation and Evaluation Details}
\label{sec:appendix}

\subsection{Closed-Form Box-Filtered Frequency-Band Encoding}
\label{supp:expected-frequency-derivation}

For query $i$ in view $v$, let $a=u_{v,i}$, $b=u_{v,i}+\Delta_v$, and $\Delta_v=b-a$.
A normalized box filter averages each sinusoidal component over this interval.
Using the antiderivatives of cosine and sine, we obtain
\begin{equation}
    \begin{aligned}
    \bar C_j
    &=\frac{1}{\Delta_v}\int_a^b\cos(2\pi f_j u)\,du
    =\left[\frac{\sin(2\pi f_j u)}{2\pi f_j\Delta_v}\right]_a^b\\
    &
    =\frac{\sin(2\pi f_jb)-\sin(2\pi f_ja)}{2\pi f_j\Delta_v},\\
    \bar S_j
    &=\frac{1}{\Delta_v}\int_a^b\sin(2\pi f_j u)\,du
    =\left[-\frac{\cos(2\pi f_j u)}{2\pi f_j\Delta_v}\right]_a^b\\
    &
    =\frac{\cos(2\pi f_ja)-\cos(2\pi f_jb)}{2\pi f_j\Delta_v}.
    \end{aligned}
    \label{eq:expected-pe}
\end{equation}

Concatenating these coefficients and dividing by $\sqrt{F}$ gives $\bar{\gamma}_{v,i}$.

\input{sections/evaluation_supplement}

%% file: sections/evaluation_supplement.tex
\subsection{Reference-View Selection Beyond Official Settings}
\label{supp:evaluation-details}

\paragraph{RealEstate10K.}
When we require more reference views beyond the official dataset settings, we retain the official reference views and incrementally add frames using farthest-point sampling in the temporal domain, which reduces redundancy among selected views.
Let $\mathcal{R}$ denote the current set of selected reference-frame indices and $\mathcal{C}$ the candidate frames from the source video, which exclude the target views and the selected reference views.
At each step, we select
\[
f^*=\arg\max_{f\in\mathcal{C}\setminus\mathcal{R}}
\min_{r\in\mathcal{R}} |f-r|,
\]
This process is repeated until the desired number of reference views is reached.

\paragraph{DL3DV.}
We use the frame indices provided by LagerNVS~\citep{szymanowicz2026lagernvs} to define each 50-frame window.
Within each 50-frame window, we first use temporal farthest-point sampling to meet the desired number of reference views beyond the official setting.
We then use the remaining frames in the window as target views, so $V$ reference views leave $50-V$ target views.

Fig.~\ref{fig:fixed-budget-additional-metrics} instead uses a fixed target set across all reference-view counts for DL3DV. 
Under the original DL3DV protocol, the target set is defined as the complement of the reference views and therefore changes with view count. To enable direct quality comparison when scaling views, we use the same 40 targets per scene from the 10-view setting for all DL3DV results in this figure. This set is a subset of the original targets at every reference-view count.

\subsection{Training Details}
\label{supp:finetuning-details}

We follow the 1--10-view training recipe of the LagerNVS general model~\citep{szymanowicz2026lagernvs}, with a few modifications.

\paragraph{LagerNVS fine-tuning.}
We fine-tune LagerNVS for 100K iterations with a 1.2K-step warmup, using learning rates of $10^{-4}$ on RealEstate10K and $10^{-5}$ on DL3DV. We follow its dynamic batching strategy with a base batch size of 128, which decreases the batch size for larger reference-view counts.

\paragraph{\proj training.}
We train \proj for 50K iterations with a 1K-step warmup, using a learning rate of $10^{-4}$ and a base batch size of 64.
For the culling-aware model, we further fine-tune for 25K iterations with a 1K-step warmup, using a learning rate of $10^{-5}$ and the same base batch size.

\subsection{Depth Denoising for Point-Cloud Visualization}
\label{supp:depth-visualization}

To visualize the noisy predicted reference-view depth as a point cloud, we apply a $3\times3$ median filter to the patch-level depth and upsample it to image resolution using joint bilateral upsampling~\citep{kopf2007joint} guided by the corresponding RGB image.
We then back-project the filtered depth using the camera intrinsics, transform the points into a shared scene coordinate system using the camera poses, and color them with RGB values.
This post-processing is used only for point-cloud visualization; the displayed depth maps are raw predictions, and the model geometry, anchors, and rendering remain unchanged.

\subsection{Rendering-Speed Evaluation}
\label{supp:speed-protocol}

\begin{figure}[t]
\centering
\includegraphics[width=\linewidth]{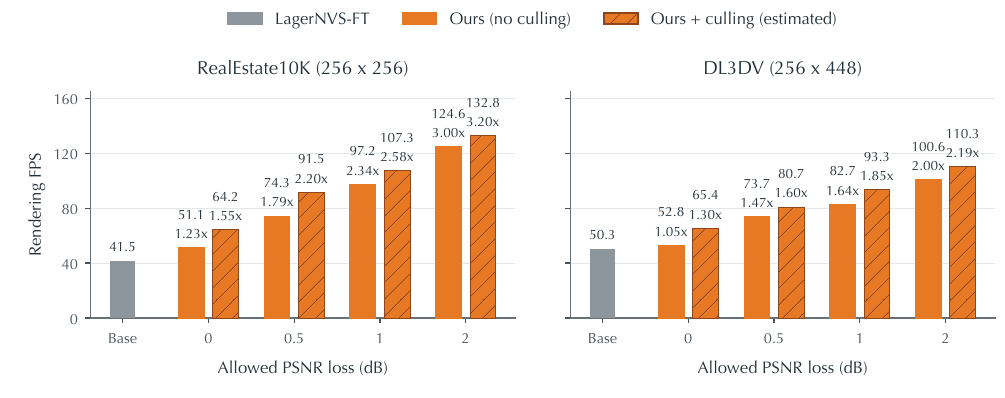}
\caption{Rendering FPS at 10 reference views on an A100 80GB with BF16.
Left: RealEstate10K; right: DL3DV.
Annotations show FPS and speedup over \mode{LagerNVS-FT}.
For \mode{Ours + Culling}, FPS is estimated by measuring rendering speed at the average post-culling decoder-token count over the test set.}
\label{fig:rendering-speed}
\end{figure}

We report results for \proj both with and without culling in \Fig{fig:rendering-speed}.
We evaluate rendering speed on an NVIDIA A100 80GB GPU using BF16 with a batch size of 1 and 1 target view.
For the baseline and our no-culling model, we directly benchmark rendering latency at fixed scene-token counts.
For the culling-enabled model, the effective decoder-token count varies across scenes and target poses.
We therefore estimate its speed by measuring rendering speed at the average post-culling decoder-token count over the test set.
The estimated rendering latency includes the measured culling overhead.

\subsection{Scene Semantic Understanding Evaluation Protocols}
\label{supp:semantic-probe} 

We first test scene-semantic understanding using VLM-generated annotations for RealEstate10K~\citep{he2024cameractrl,li2023lavis} to categorize scenes into eight semantic classes: living room, bedroom, kitchen, bathroom, dining room, hallway, staircase, and outdoor.
Each scene receives one class derived from its caption.
We filter out scenes that belong to multiple classes or do not belong to any of the predefined classes. This results in 40,857 training scenes and 4,606 validation scenes.
We train for 500 epochs with a learning rate of $3\times10^{-2}$, batch size 256, and weight decay $10^{-4}$. We use cosine annealing to decay the learning rate to zero over training.

\begin{figure}[t]
\centering
\includegraphics[width=0.49\linewidth]{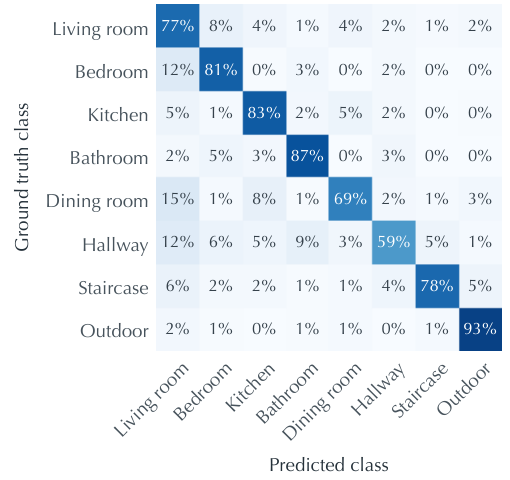}
\caption{Row-normalized confusion (\%) for our scene classification.
Top-1/2 accuracy: 80.40/91.73\% on 4,606 validation scenes.}
\label{fig:semantic-readout}
\label{fig:semantic-probe}
\end{figure}

\section{More Results}

\paragraph{Our Base Model is Strong}
\label{supp:base-model-comparison}

We provide more results for \Sect{sec:base_model_comp}, including PSNR, SSIM, and LPIPS comparisons with 2--16 reference views on both datasets (\Fig{fig:base-model-additional-metrics}).

\paragraph{\proj Achieves the Best Quality--Token-Count Trade-offs}
\label{supp:system-results}

We provide more results for \Sect{sec:base_model_comp}, including PSNR, SSIM, and LPIPS versus scene-token counts for \proj and \mode{CLiFT} with 2, 4, 6, and 10 reference views on both datasets (\Fig{fig:system-additional-metrics}).

\paragraph{Comparing against Different Condenser Designs}
\label{supp:view-scaling}

We provide more results for \Sect{sec:compression-results}, including PSNR, SSIM, and LPIPS comparisons across condenser designs with 2, 4, 6, and 10 reference views on both datasets (Figs.~\ref{fig:condenser-additional-metrics}--\ref{fig:condenser-additional-10views}).
All methods use the same base model with culling disabled.

\paragraph{Scene-Token Reductions \& Rendering Speedup at Fixed Quality Target.}
\label{supp:rendering-speed}

We provide more results for \Sect{sec:quality-budget-savings}, including  rendering FPS and speedups at 0, 0.5, 1, and 2\,dB quality-loss budgets on both datasets with 10 reference views (\Fig{fig:rendering-speed}).
The evaluation protocol is described in \ref{supp:speed-protocol}.

\paragraph{Scaling Quality with Reference Views and Token Budget}
\label{supp:quality-scaling}
\label{supp:fixed-budget-scaling}
\label{supp:beyond-input-budget}

We provide more results for \Sect{sec:fixed-budget-scaling}, including PSNR, SSIM, and LPIPS on both datasets for increasing reference-view counts at fixed budgets of 512, 1,024, and 2,048 tokens (\Fig{fig:fixed-budget-both-datasets}) and increasing anchor budgets from 100\% to 150\% of the input token count (\Fig{fig:beyond-input-additional-metrics}).
Culling is disabled in both experiments.
For DL3DV with two reference views, the 2,048-anchor setting exceeds the corresponding highway token count of 1554.
This is enabled by our continuous anchor field (\Sect{sec:continuous-anchors}), which can sample an arbitrary number of anchors beyond the input token count.

\paragraph{Ablation Studies}
\label{supp:ablation-results}

We provide more results for \Sect{sec:culling-evaluation}, including PSNR, SSIM, and LPIPS for the same component ablations with 2, 4, 6, and 10 reference views on DL3DV (Figs.~\ref{fig:ablation-supp-2-4views}--\ref{fig:ablation-supp-6-10views}).

\paragraph{\proj as a General Scene Representation}
\label{supp:beyond-rendering}

We provide  scene-classification confusion matrix (\Fig{fig:semantic-probe}) for \Sect{sec:beyond-rendering}.
The downstream evaluation protocols are given in \ref{supp:semantic-probe}.

\input{sections/compression_visuals}

\paragraph{Qualitative Analysis of Emergent Geometry and Anchor Allocation}
\label{supp:qualitative-evaluation}

We provide more results for \Sect{sec:qualitative-analysis}, including emergent depth, point-cloud, 3D anchors, and projected-anchor visualizations for 12 RealEstate10K scenes and 15 DL3DV scenes using four reference views and 2,048 anchors (Figs.~\ref{fig:qualitative-re10k-1}--\ref{fig:qualitative-dl3dv-3}).

\AtEndDocument{%
  \clearpage
  \input{sections/supplementary_figures}}

%% file: sections/compression_visuals.tex
\paragraph{Visual Comparison across Different Condensers.}
\label{supp:compression-visuals}
We provide more results for \Sect{sec:compression-results}, including 34 selected novel-view examples with 2 and 10 reference views at 0\%, 20\%, 40\%, 60\%, and 80\% token reduction on both datasets (Figs.~\ref{fig:compression-visual-07}--\ref{fig:compression-visual-06}).
For compact figure files, we JPEG-compress the full images at quality 88 without chroma subsampling (4:4:4), while keeping the zoom-in crops losslessly encoded.

%% file: sections/supplementary_figures.tex
\begingroup
\makeatletter
\setlength{\@fptop}{0pt}
\setlength{\@fpsep}{12pt plus 1fil}
\makeatother
\begin{figure}[p]
\centering
\includegraphics[width=\linewidth]{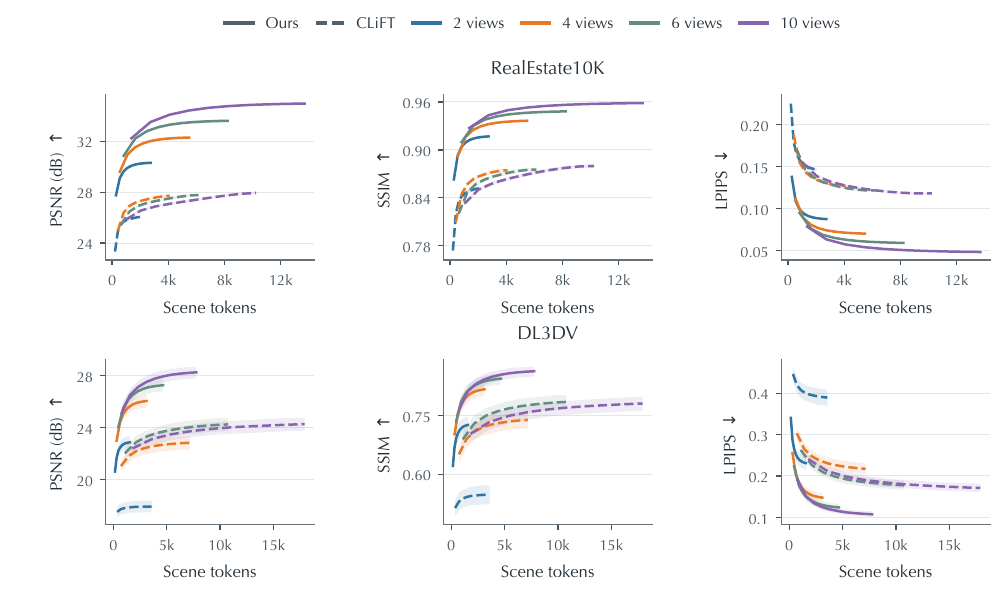}
\caption{System-level quality versus scene tokens.
Top: RealEstate10K; bottom: DL3DV. Columns show PSNR, SSIM, and LPIPS.
Colors distinguish reference-view counts; solid lines denote \proj and dashed lines denote official CLiFT.
Shaded bands show 95\% CIs.}
\label{fig:system-additional-metrics}
\end{figure}

\begin{figure}[p]
\centering
\includegraphics[width=\linewidth]{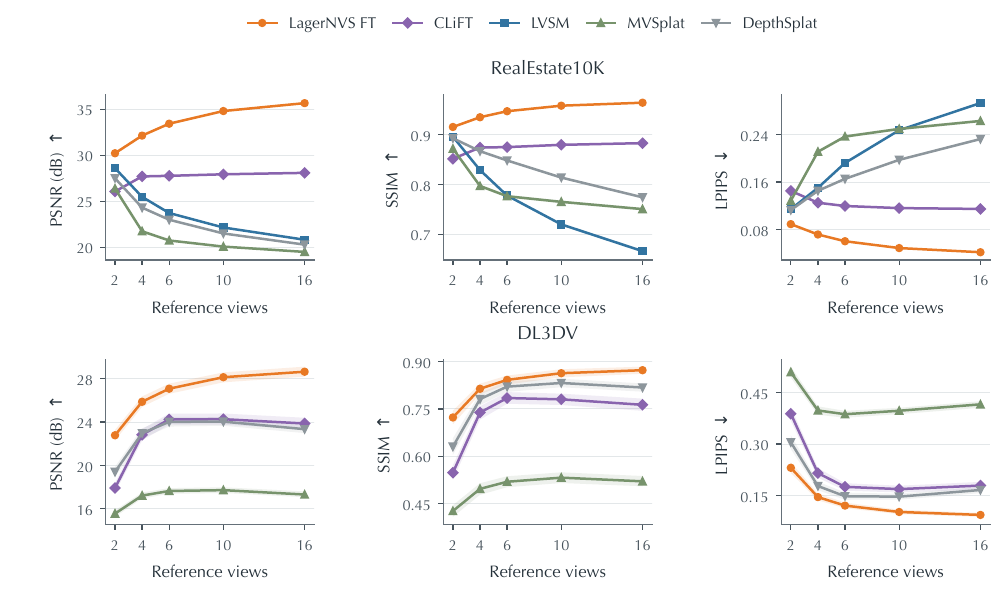}
\caption{Base-model comparison without our condenser.
Top: RealEstate10K; bottom: DL3DV. Columns show PSNR, SSIM, and LPIPS.
Shaded bands show 95\% CIs.
No official DL3DV checkpoint is available for LVSM.}
\label{fig:base-model-additional-metrics}
\end{figure}

\begin{figure}[p]
\centering
\includegraphics[width=\linewidth]{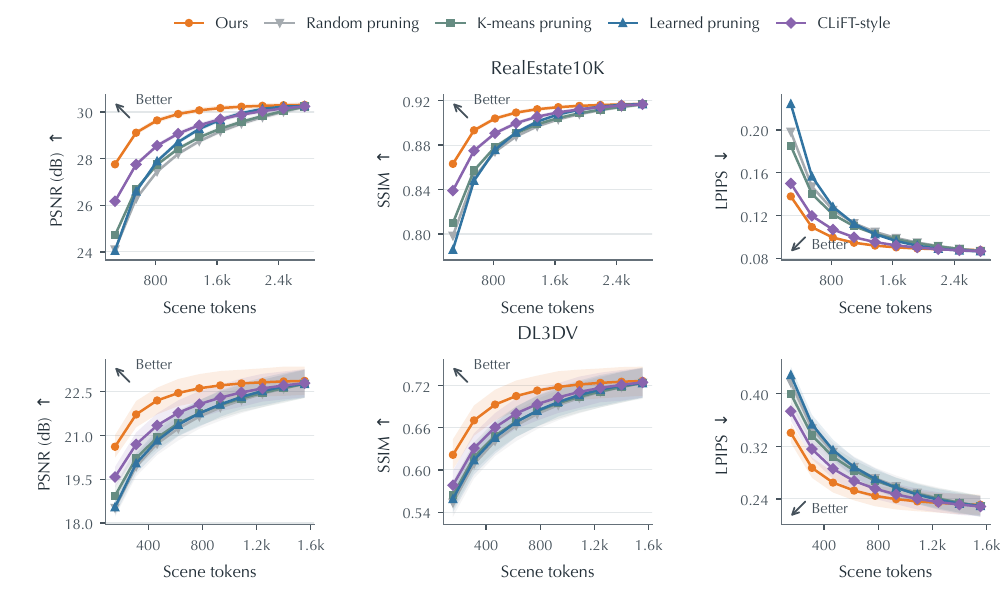}
\caption{Condenser quality--token trade-offs with 2 reference views and culling disabled.
Top: RealEstate10K; bottom: DL3DV. Columns show PSNR, SSIM, and LPIPS.
Shaded bands show 95\% CIs.}
\label{fig:condenser-additional-metrics}
\end{figure}

\begin{figure}[p]
\centering
\includegraphics[width=\linewidth]{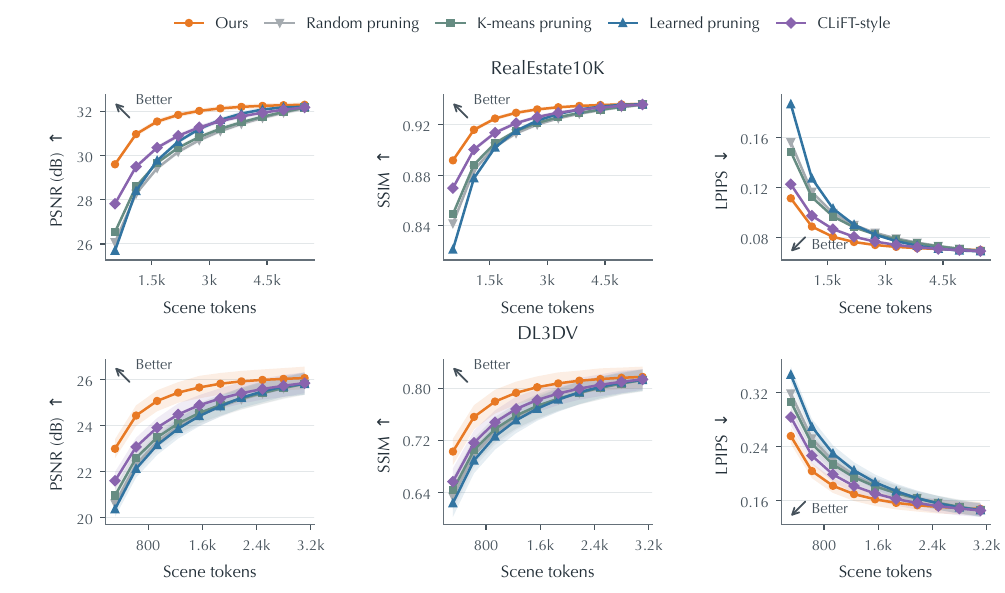}
\caption{Condenser quality--token trade-offs with 4 reference views and culling disabled.
Top: RealEstate10K; bottom: DL3DV. Columns show PSNR, SSIM, and LPIPS.
Shaded bands show 95\% CIs.}
\label{fig:condenser-additional-4views}
\end{figure}

\begin{figure}[p]
\centering
\includegraphics[width=\linewidth]{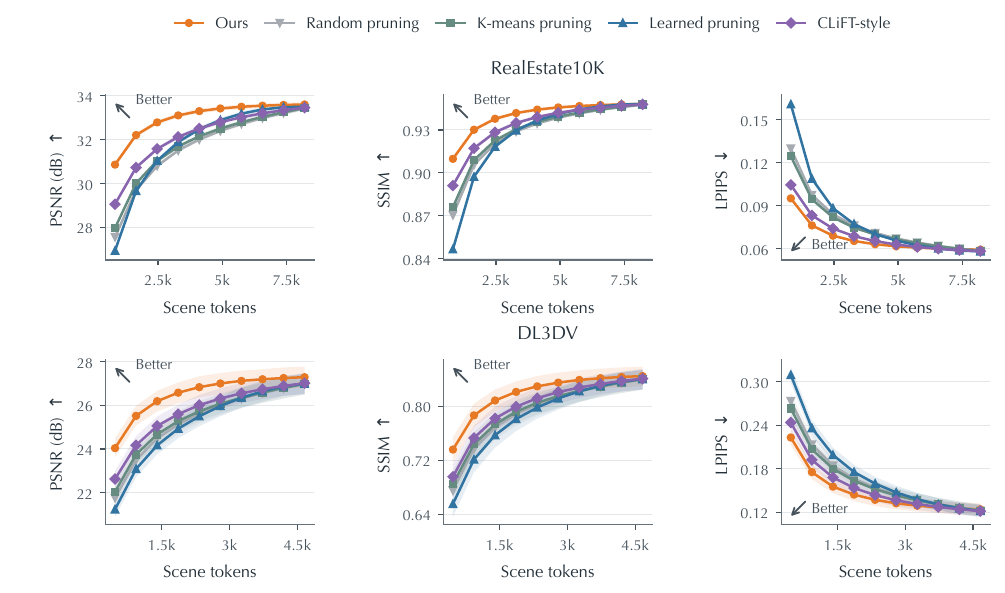}
\caption{Condenser quality--token trade-offs with 6 reference views and culling disabled.
Top: RealEstate10K; bottom: DL3DV. Columns show PSNR, SSIM, and LPIPS.
Shaded bands show 95\% CIs.}
\label{fig:condenser-additional-6views}
\end{figure}

\begin{figure}[p]
\centering
\includegraphics[width=\linewidth]{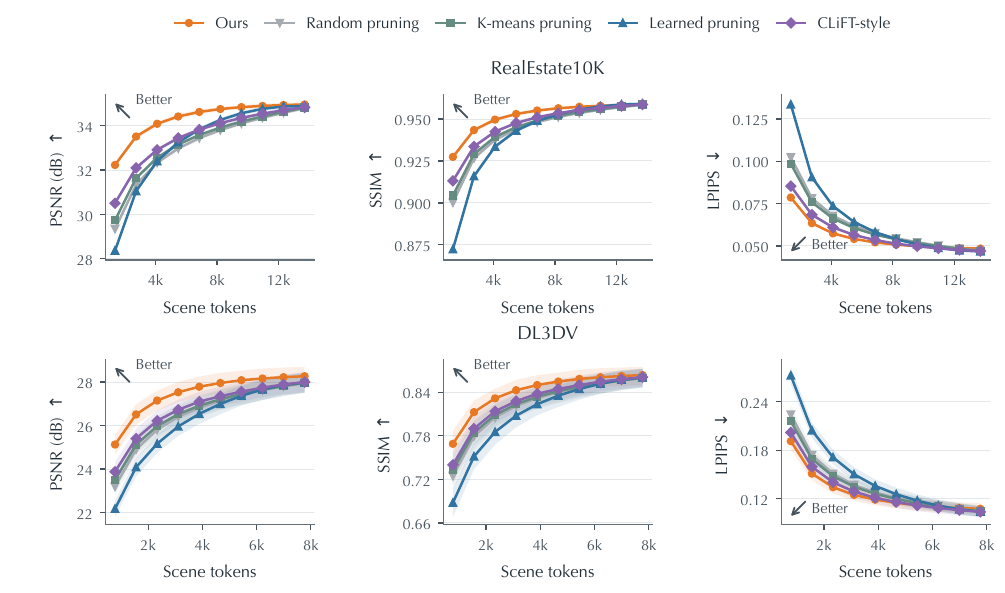}
\caption{Condenser quality--token trade-offs with 10 reference views and culling disabled.
Top: RealEstate10K; bottom: DL3DV. Columns show PSNR, SSIM, and LPIPS.
Shaded bands show 95\% CIs.}
\label{fig:condenser-additional-10views}
\end{figure}

\begin{figure}[p]
\centering
\begin{minipage}{\linewidth}
\centering
\includegraphics[width=\linewidth]{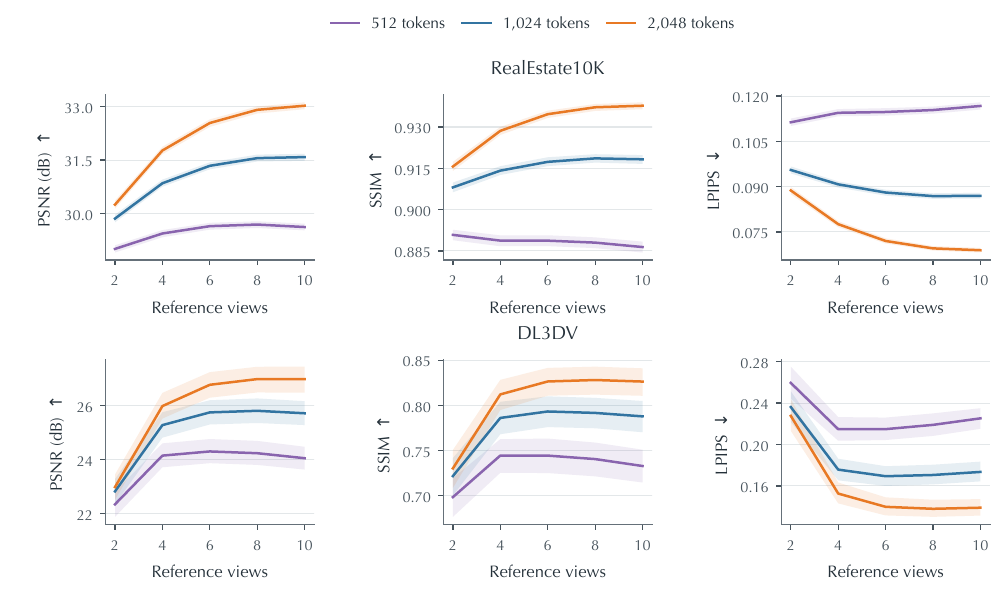}
\caption{Fixed-budget view scaling with culling disabled.
Top: RealEstate10K; bottom: DL3DV. Columns show PSNR, SSIM, and LPIPS.
Shaded bands show 95\% scene-bootstrap CIs.\textsuperscript{\emph{a}}}
\label{fig:fixed-budget-both-datasets}
\label{fig:fixed-budget-additional-metrics}
\footnotetext[1]{Under the original DL3DV protocol, the target set is the complement of the reference views and therefore changes with view count. To make scaling directly comparable, all DL3DV results in this figure use the same 40 targets per scene from the 10-view setting, a subset of the original target set at every reference-view count.}
\end{minipage}
\end{figure}

\begin{figure}[p]
\centering
\includegraphics[width=\linewidth]{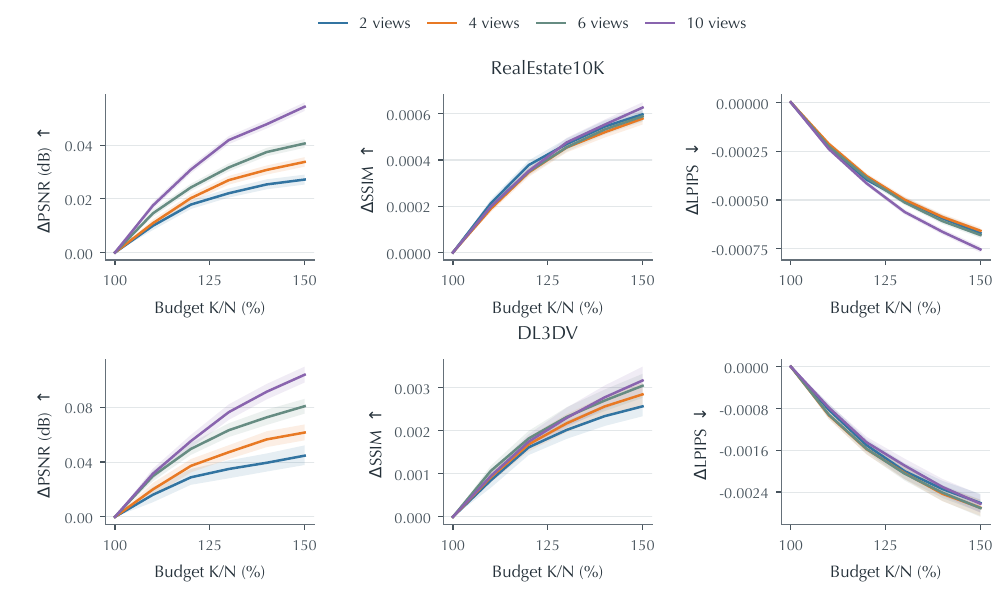}
\caption[Paired changes relative to $K=N$ with culling disabled.]{Paired changes relative to $K=N$ with culling disabled.
Top: RealEstate10K; bottom: DL3DV. Columns show $\Delta$PSNR, $\Delta$SSIM, and $\Delta$LPIPS.
Shaded bands show paired 95\% scene-bootstrap intervals.
Negative $\Delta$LPIPS indicates improvement.}
\label{fig:beyond-input-additional-metrics}
\end{figure}

\begin{figure}[p]
\centering
\includegraphics[width=\linewidth]{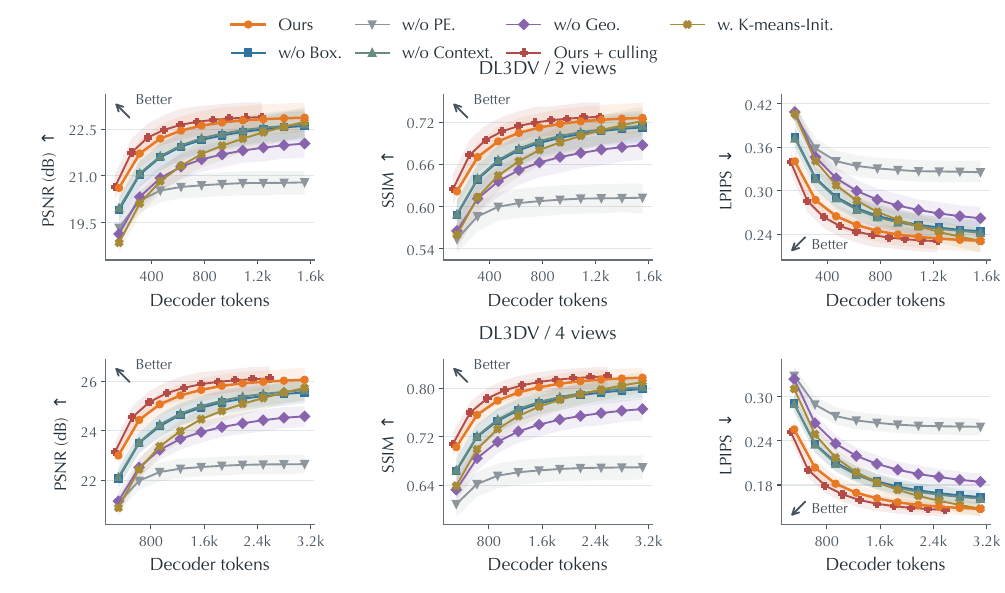}
\caption{DL3DV ablations with 2 reference views (top) and 4 reference views (bottom).
Columns show PSNR, SSIM, and LPIPS.
Shaded bands show 95\% scene-bootstrap CIs.}
\label{fig:ablation-supp-2-4views}
\end{figure}

\begin{figure}[p]
\centering
\includegraphics[width=\linewidth]{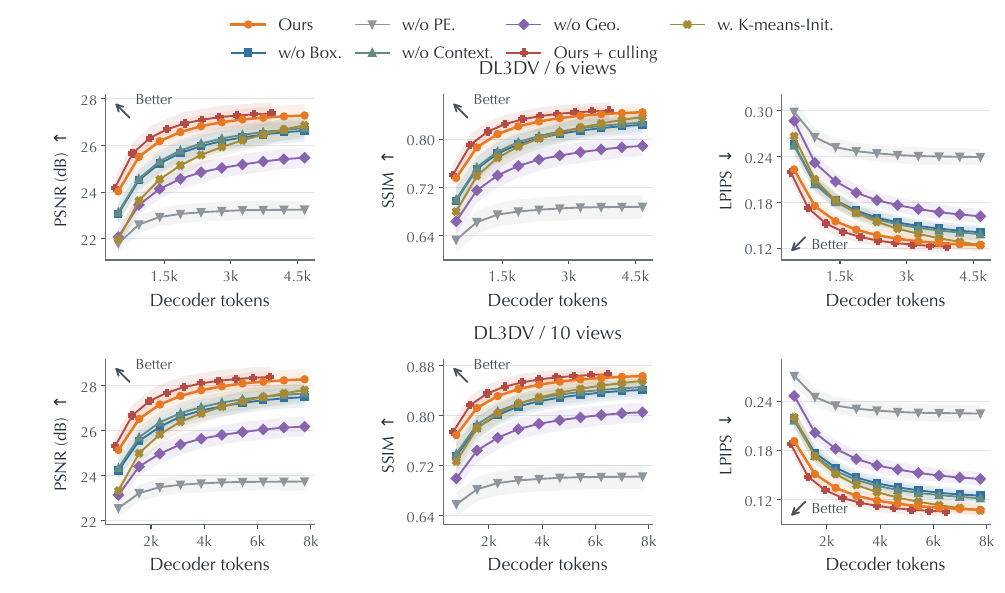}
\caption{DL3DV ablations with 6 reference views (top) and 10 reference views (bottom).
Columns show PSNR, SSIM, and LPIPS.
Shaded bands show 95\% scene-bootstrap CIs.}
\label{fig:ablation-supp-6-10views}
\end{figure}

\clearpage
\input{sections/compression_visual_figures}
\input{sections/qualitative_selected}
\clearpage
\endgroup

%% file: sections/compression_visual_figures.tex
\clearpage
\begin{figure}[p]
\centering
\includegraphics[width=\linewidth,height=0.95\textheight,keepaspectratio]{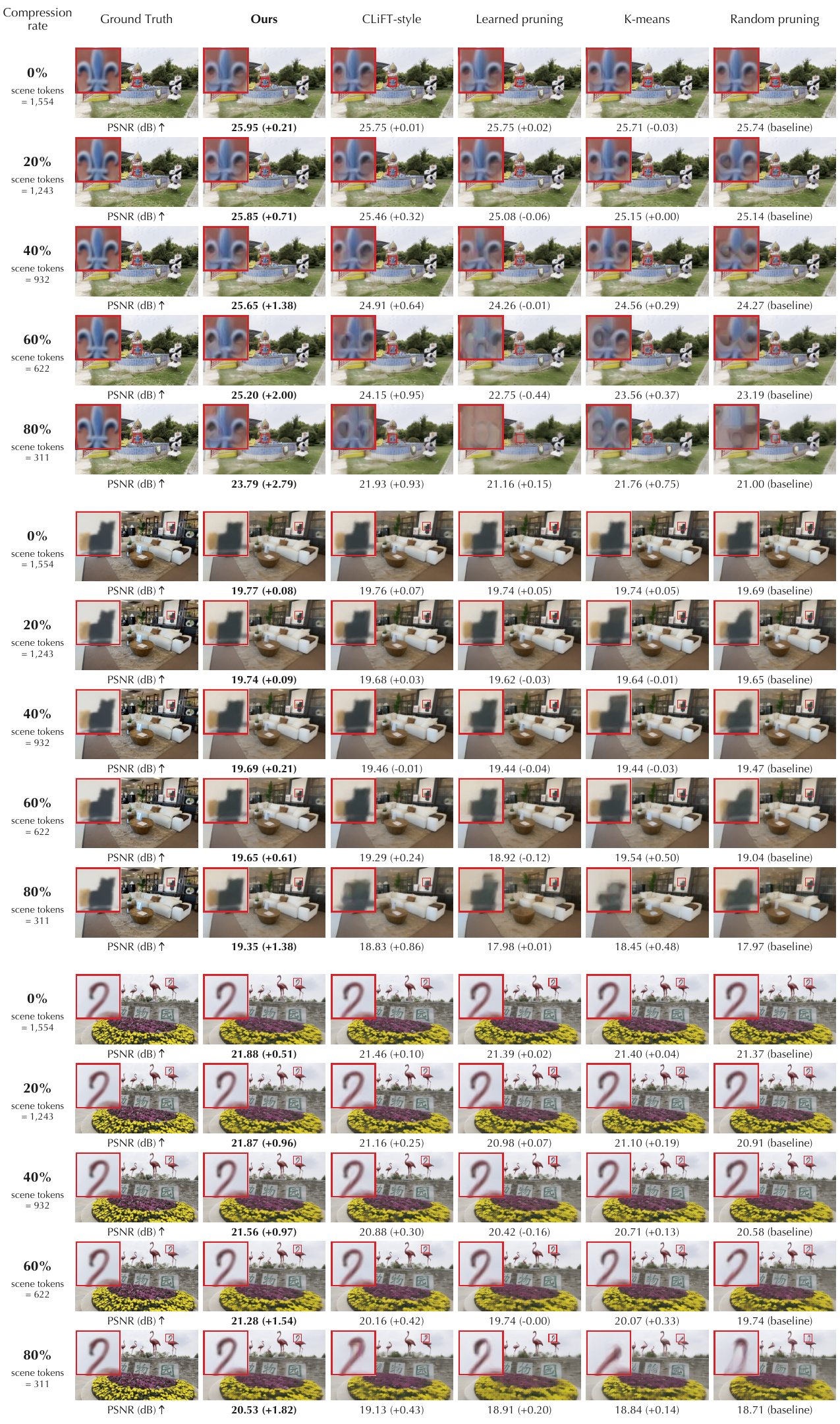}
\caption{Visual comparison of different condensers on DL3DV.}
\label{fig:compression-visual-07}
\end{figure}

\clearpage
\begin{figure}[p]
\centering
\includegraphics[width=\linewidth,height=0.95\textheight,keepaspectratio]{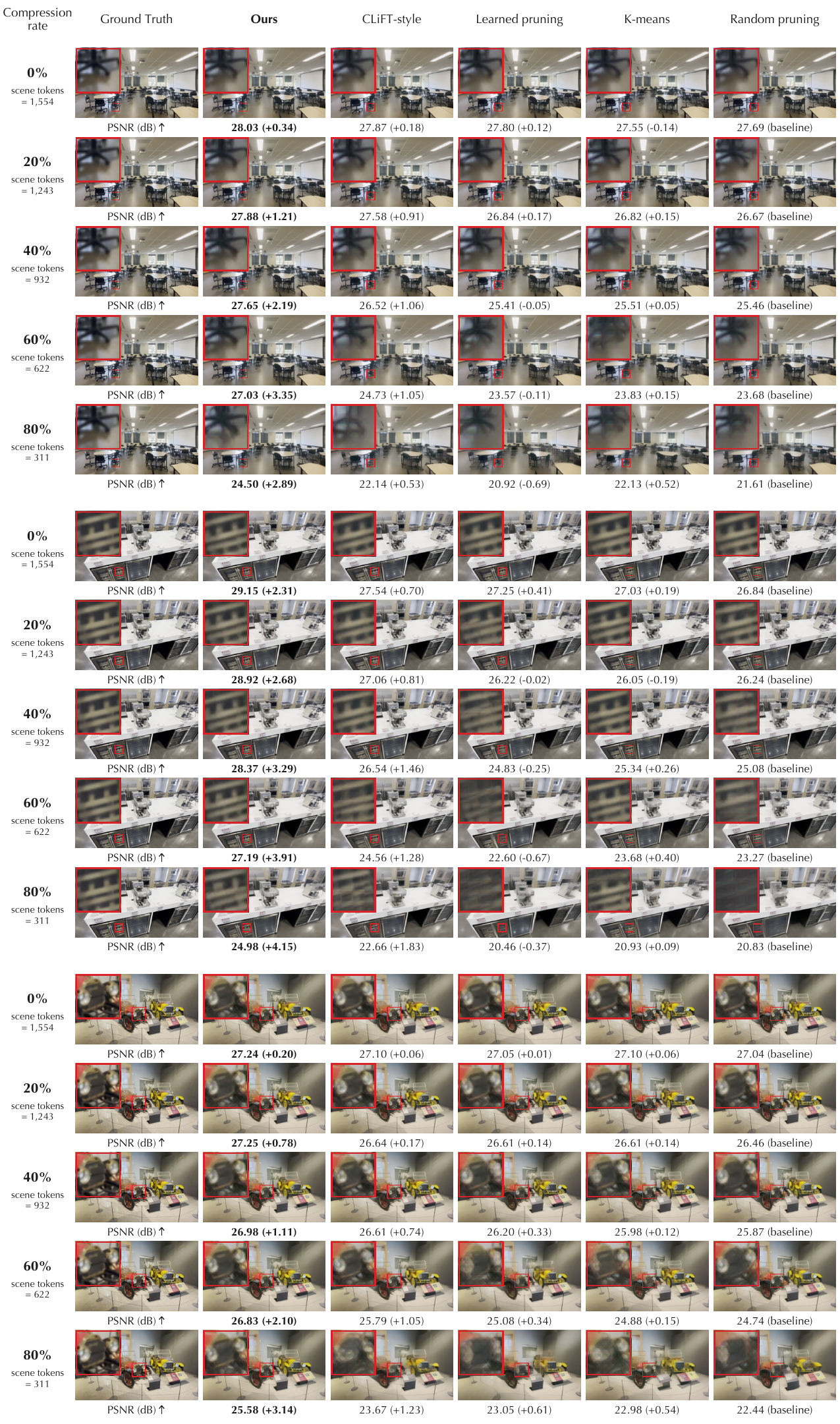}
\caption{Visual comparison of different condensers on DL3DV.}
\label{fig:compression-visual-08}
\end{figure}

\clearpage
\begin{figure}[p]
\centering
\includegraphics[width=\linewidth,height=0.95\textheight,keepaspectratio]{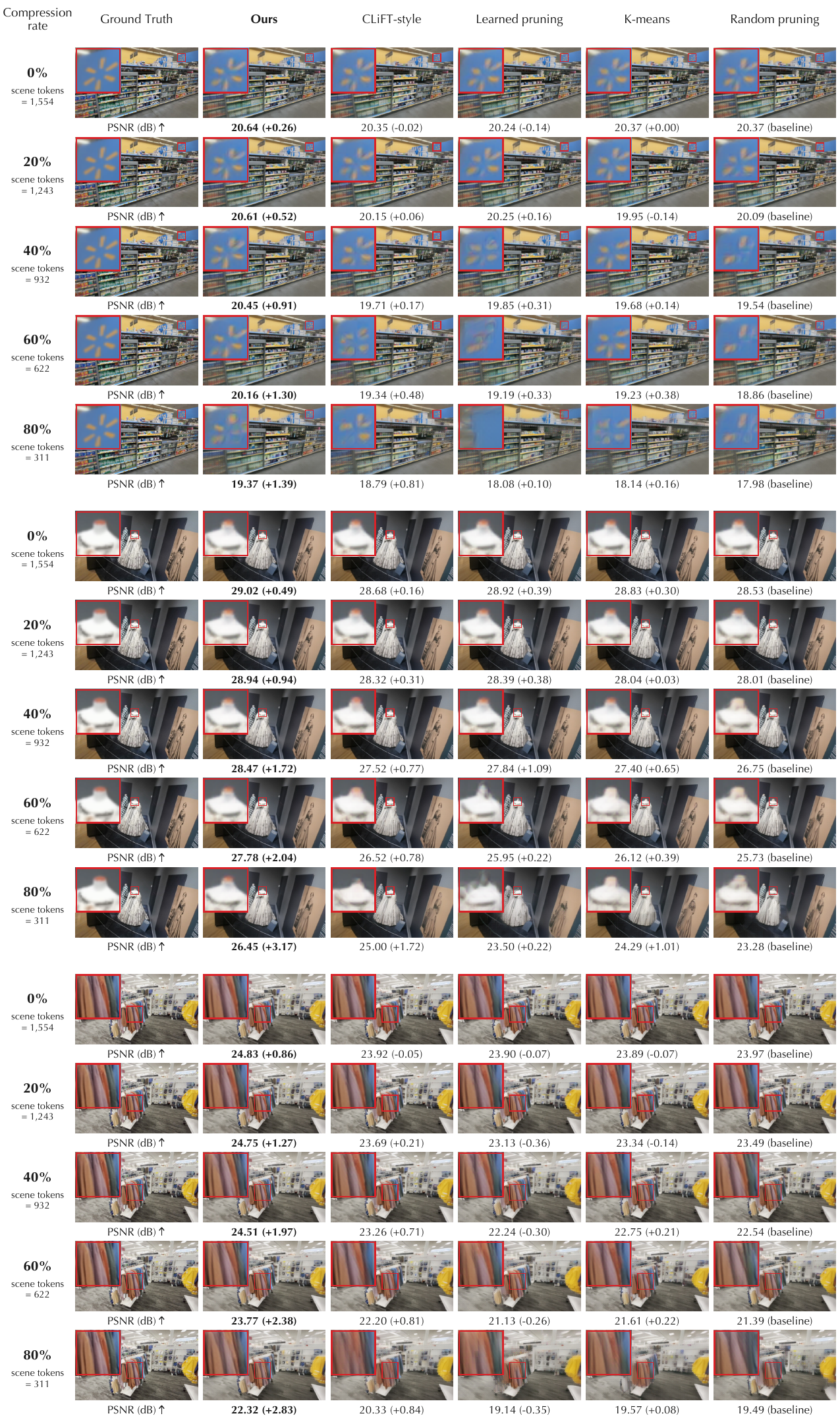}
\caption{Visual comparison of different condensers on DL3DV.}
\label{fig:compression-visual-09}
\end{figure}

\clearpage
\begin{figure}[p]
\centering
\includegraphics[width=\linewidth,height=0.95\textheight,keepaspectratio]{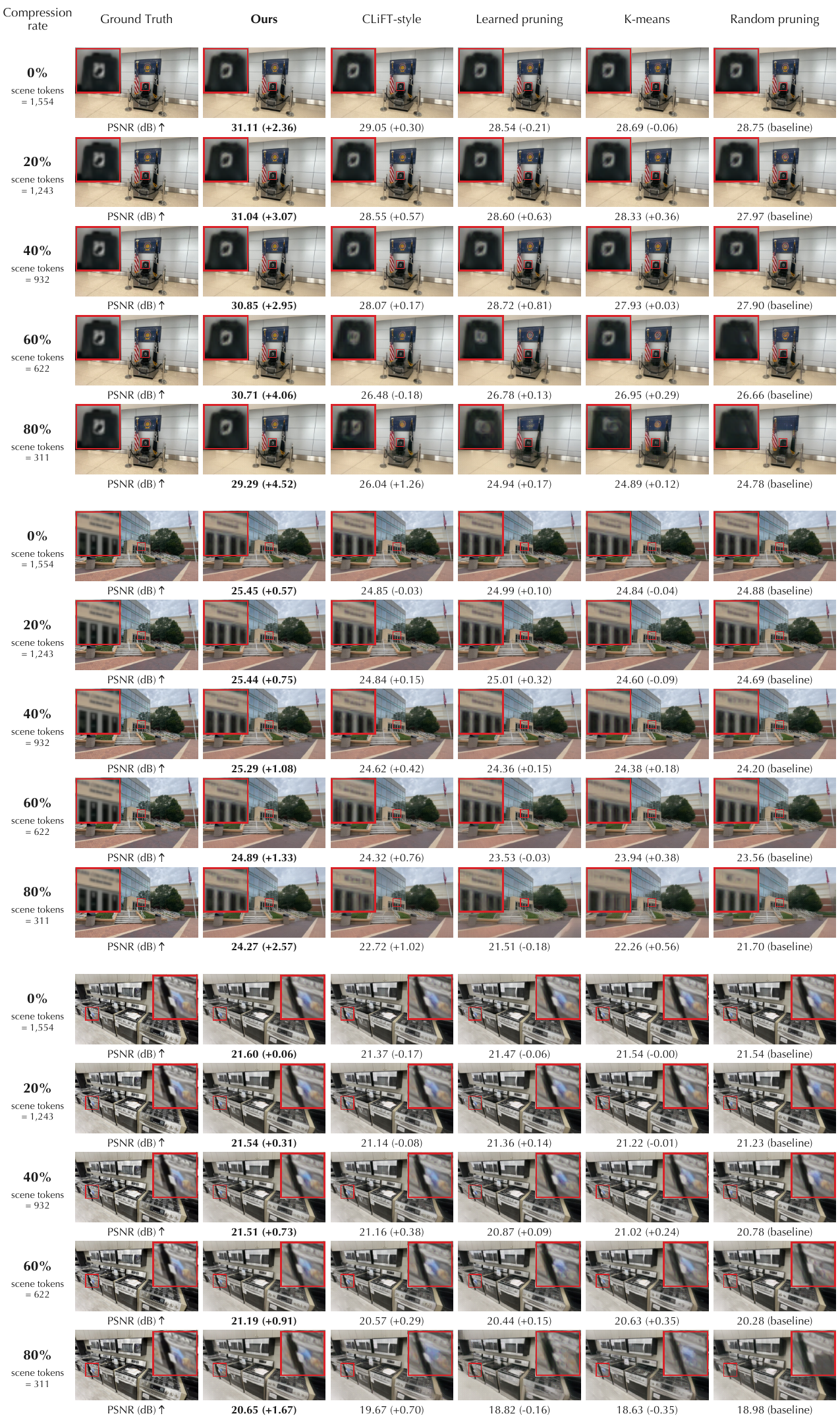}
\caption{Visual comparison of different condensers on DL3DV.}
\label{fig:compression-visual-10}
\end{figure}

\clearpage
\begin{figure}[p]
\centering
\includegraphics[width=\linewidth,height=0.95\textheight,keepaspectratio]{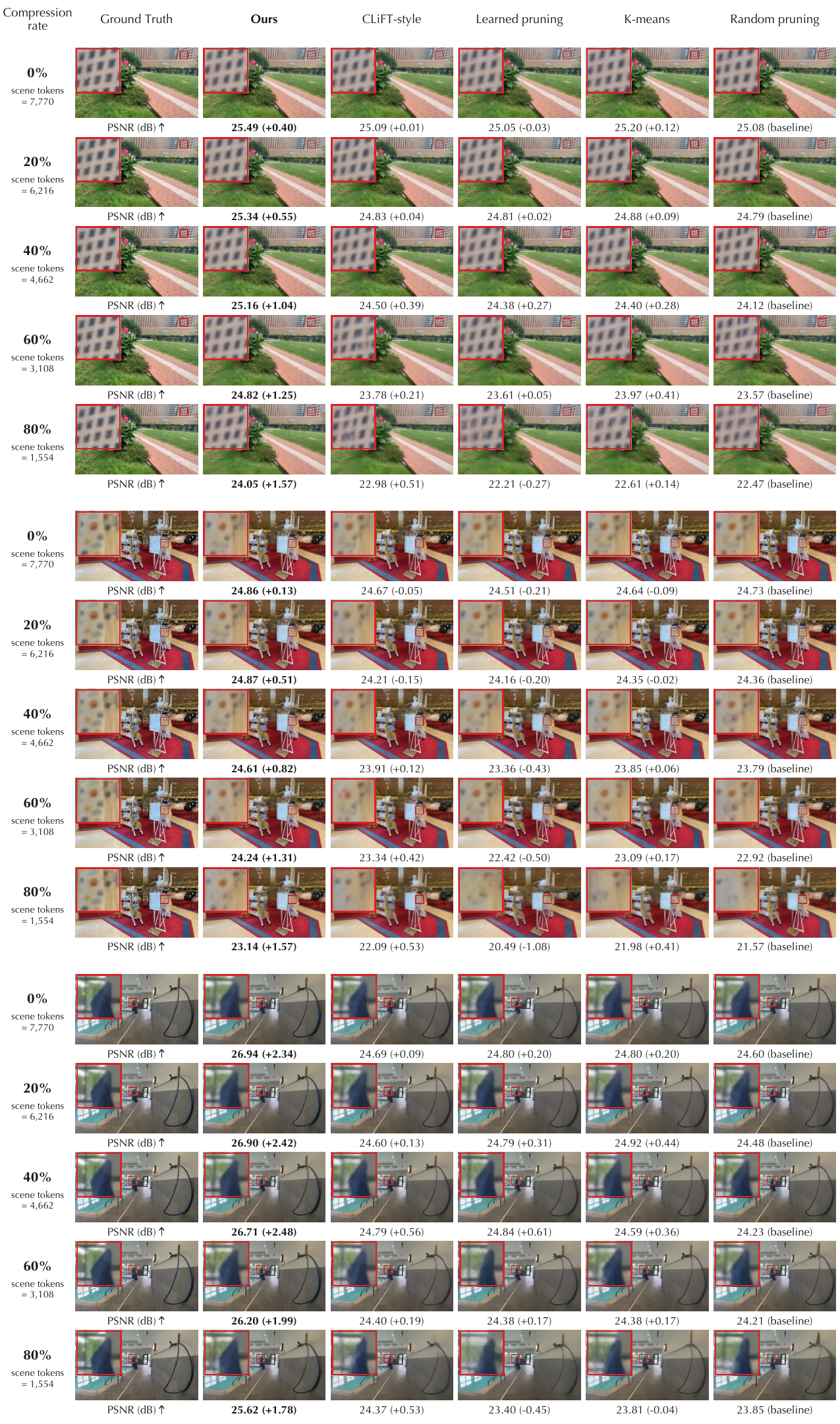}
\caption{Visual comparison of different condensers on DL3DV.}
\label{fig:compression-visual-11}
\end{figure}

\clearpage
\begin{figure}[p]
\centering
\includegraphics[width=\linewidth,height=0.95\textheight,keepaspectratio]{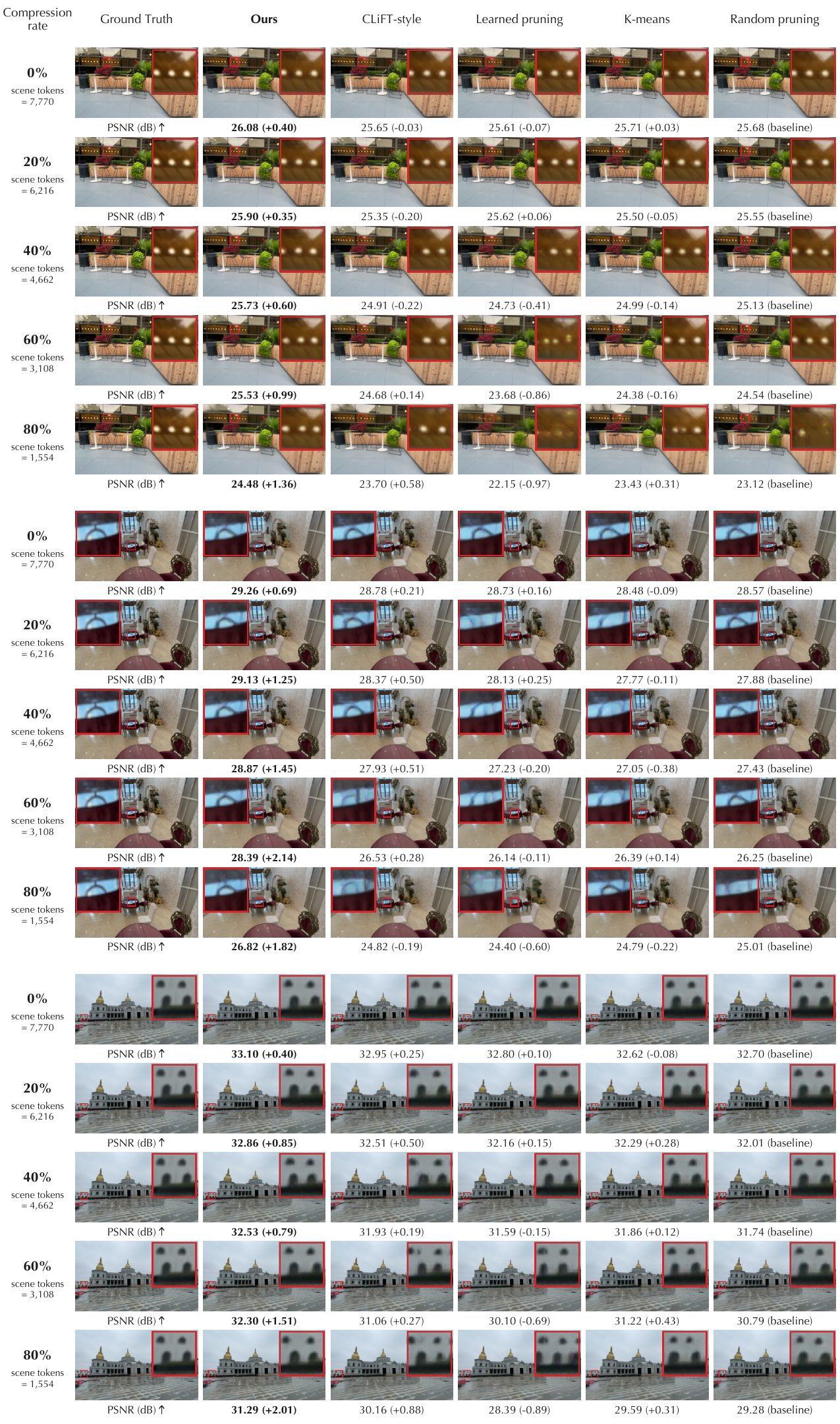}
\caption{Visual comparison of different condensers on DL3DV.}
\label{fig:compression-visual-12}
\end{figure}

\clearpage
\begin{figure}[p]
\centering
\includegraphics[width=\linewidth,height=0.95\textheight,keepaspectratio]{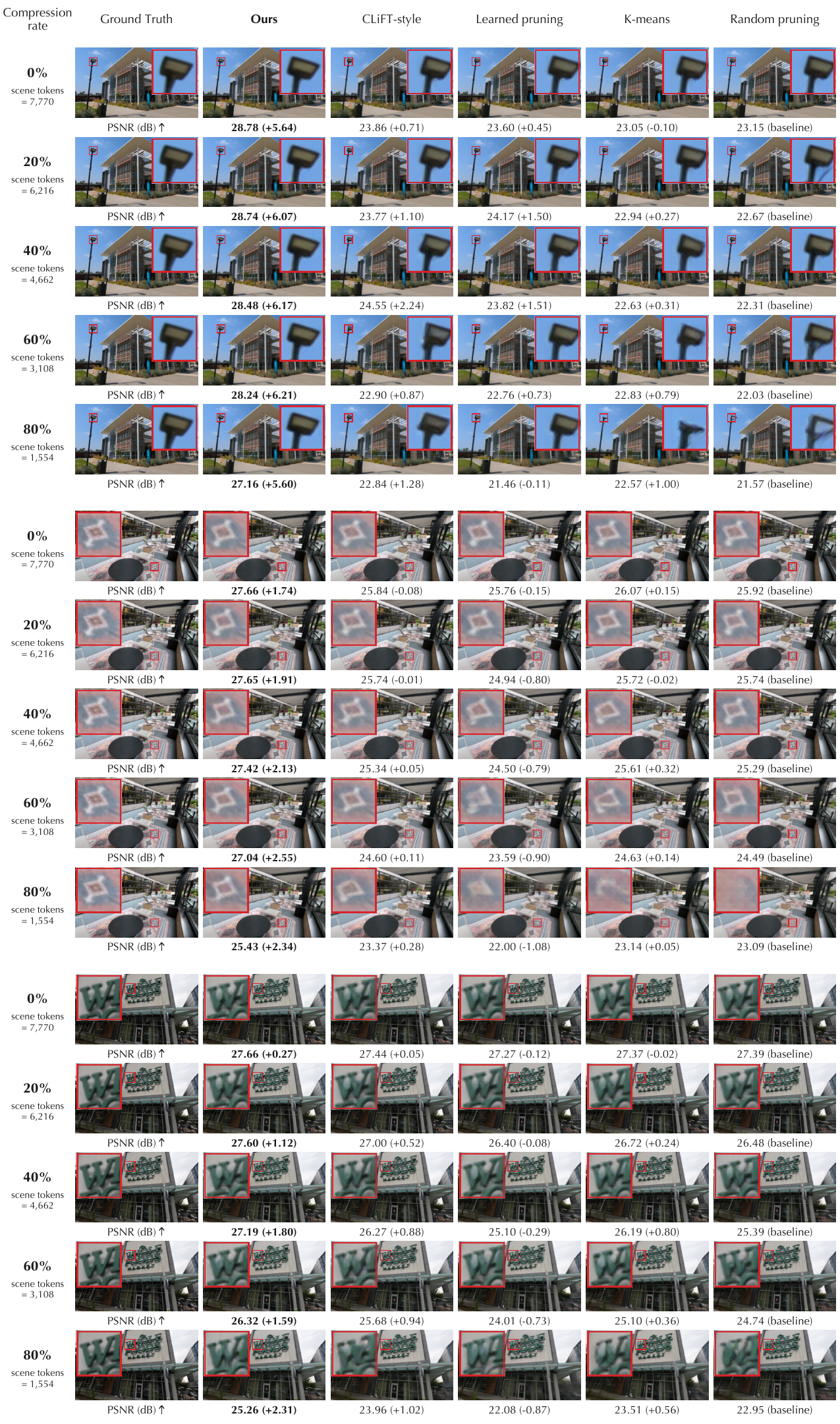}
\caption{Visual comparison of different condensers on DL3DV.}
\label{fig:compression-visual-13}
\end{figure}

\clearpage
\begin{figure}[p]
\centering
\includegraphics[width=\linewidth,height=0.95\textheight,keepaspectratio]{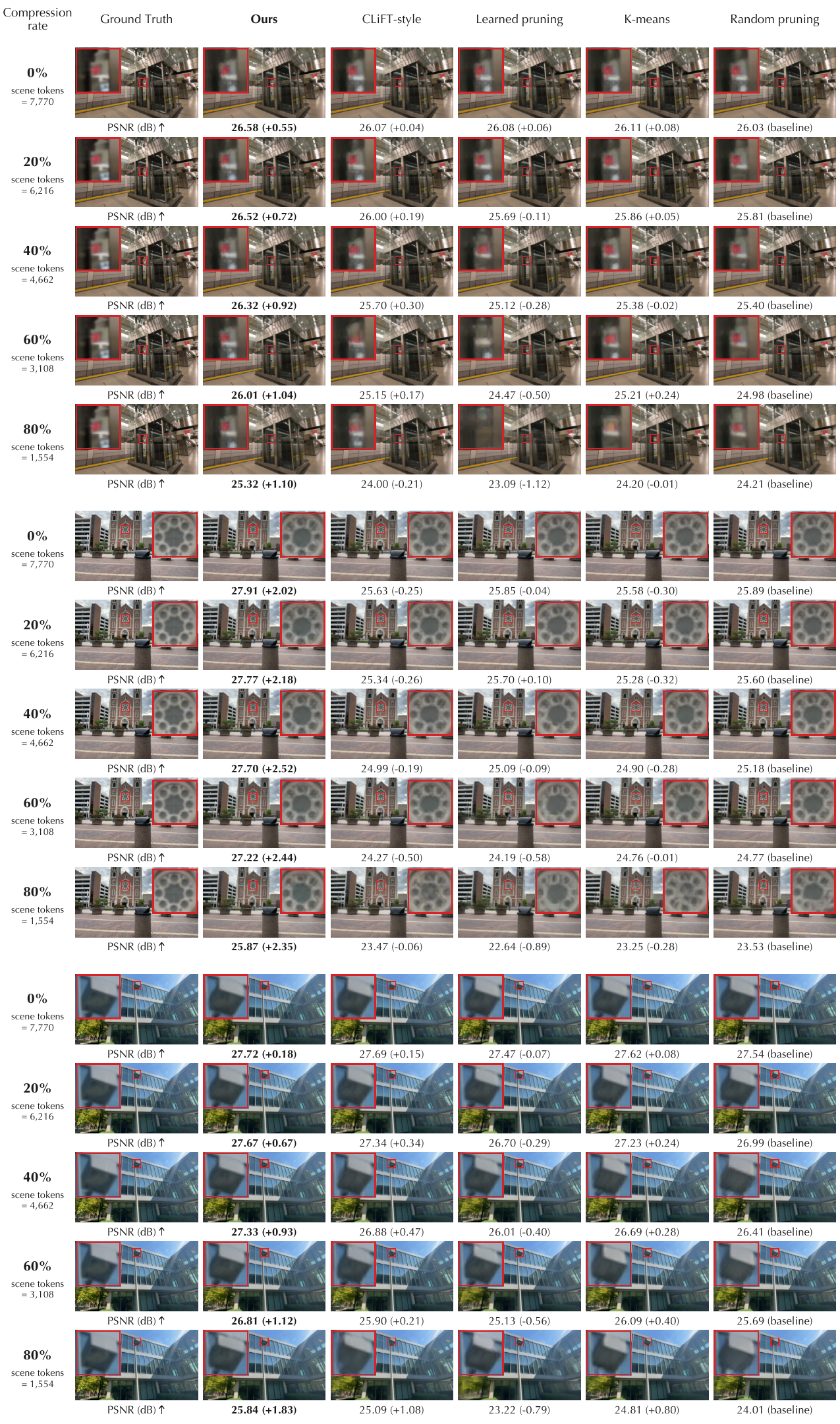}
\caption{Visual comparison of different condensers on DL3DV.}
\label{fig:compression-visual-14}
\end{figure}

\clearpage
\begin{figure}[p]
\centering
\includegraphics[width=\linewidth,height=0.95\textheight,keepaspectratio]{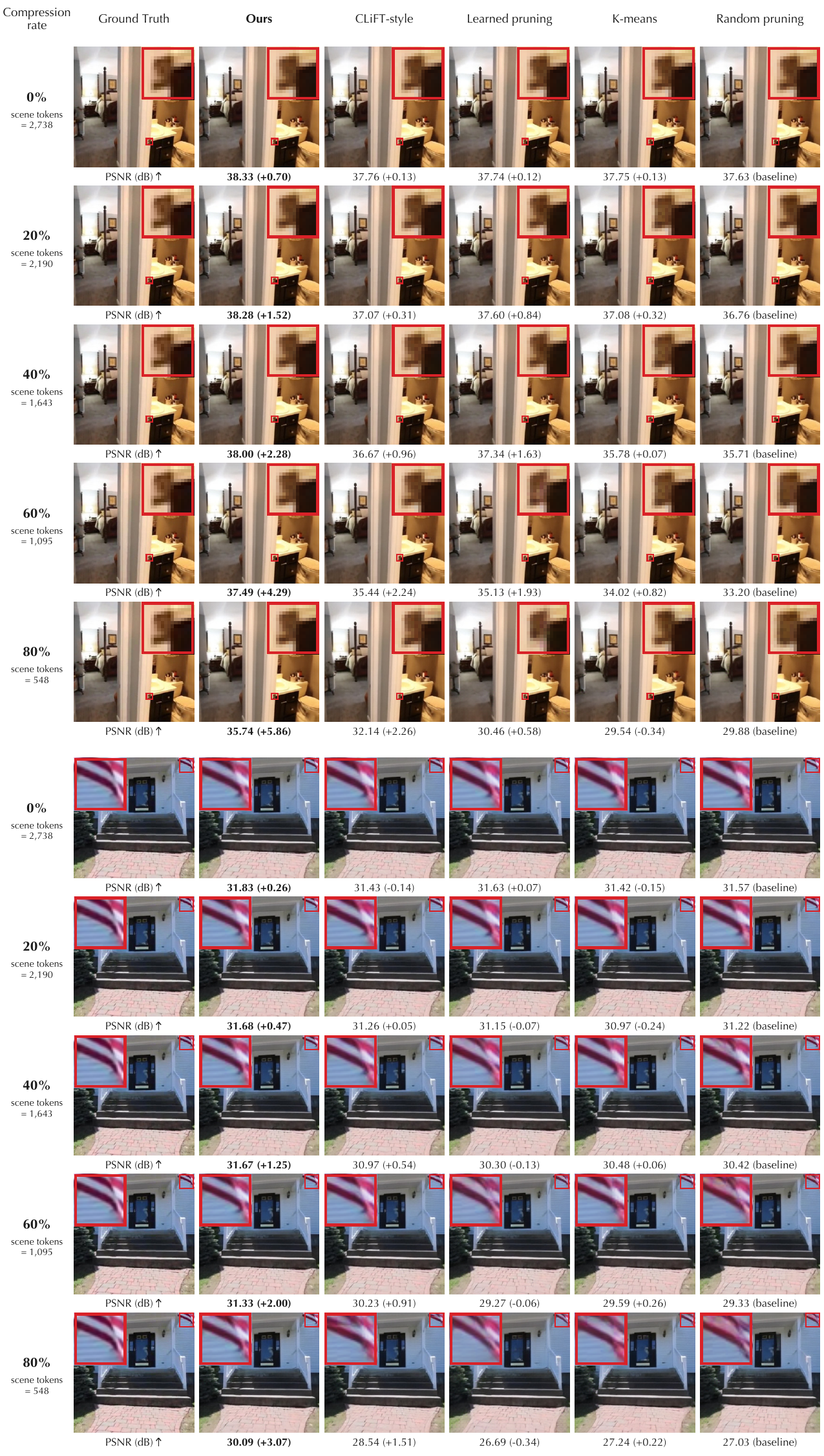}
\caption{Visual comparison of different condensers on RealEstate10K.}
\label{fig:compression-visual-01}
\end{figure}

\clearpage
\begin{figure}[p]
\centering
\includegraphics[width=\linewidth,height=0.95\textheight,keepaspectratio]{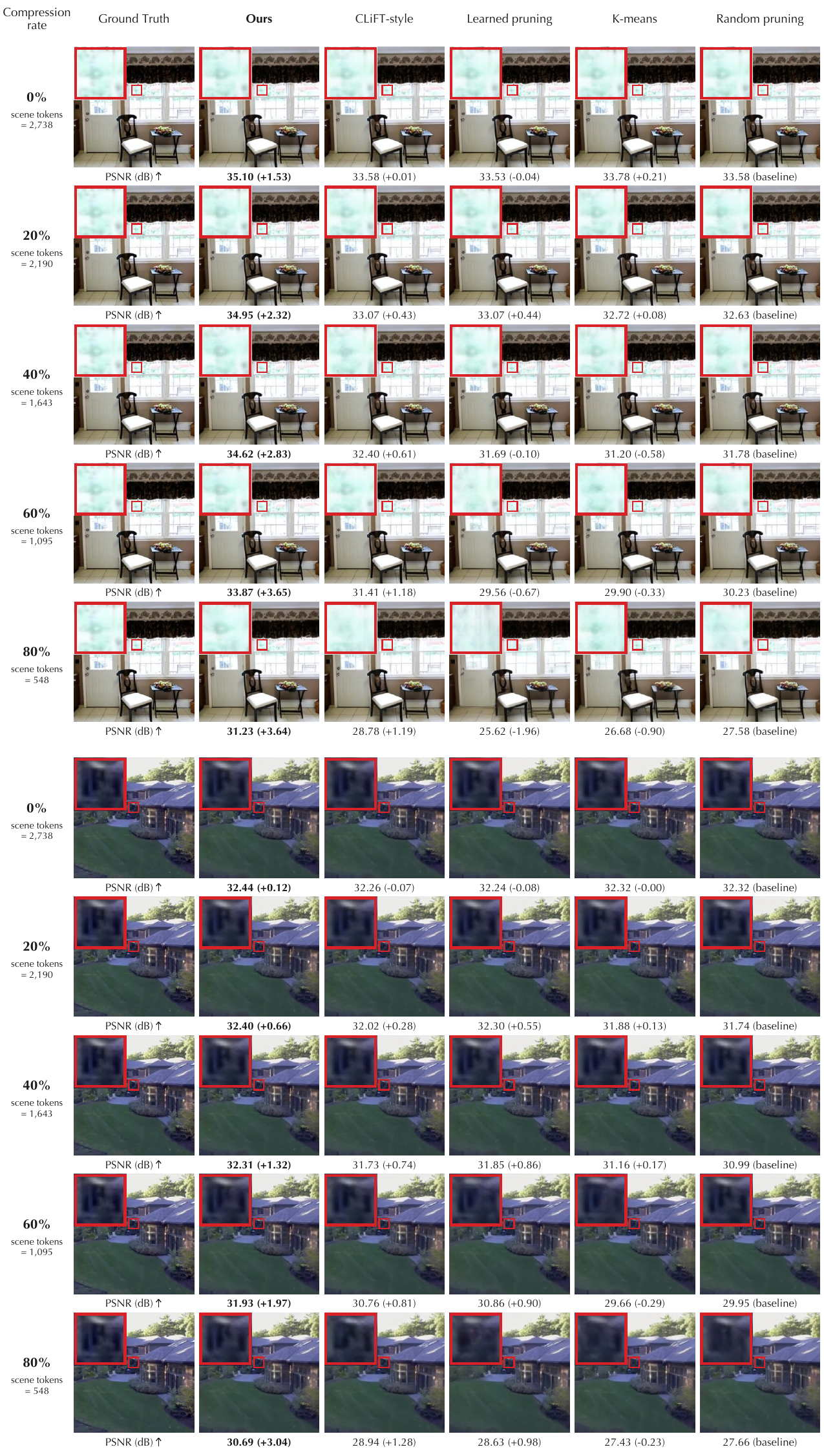}
\caption{Visual comparison of different condensers on RealEstate10K.}
\label{fig:compression-visual-02}
\end{figure}

\clearpage
\begin{figure}[p]
\centering
\includegraphics[width=\linewidth,height=0.95\textheight,keepaspectratio]{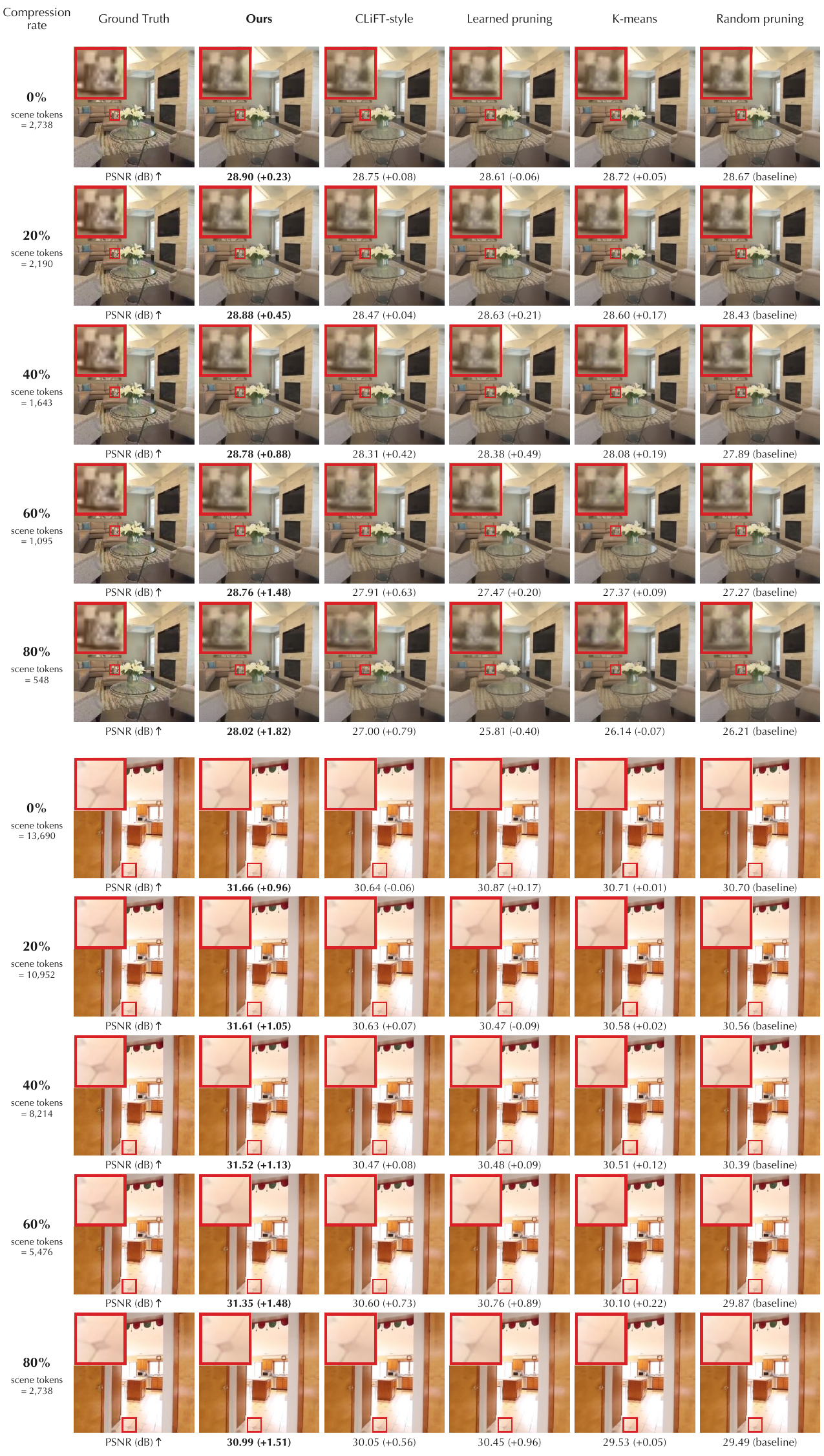}
\caption{Visual comparison of different condensers on RealEstate10K.}
\label{fig:compression-visual-03}
\end{figure}

\clearpage
\begin{figure}[p]
\centering
\includegraphics[width=\linewidth,height=0.95\textheight,keepaspectratio]{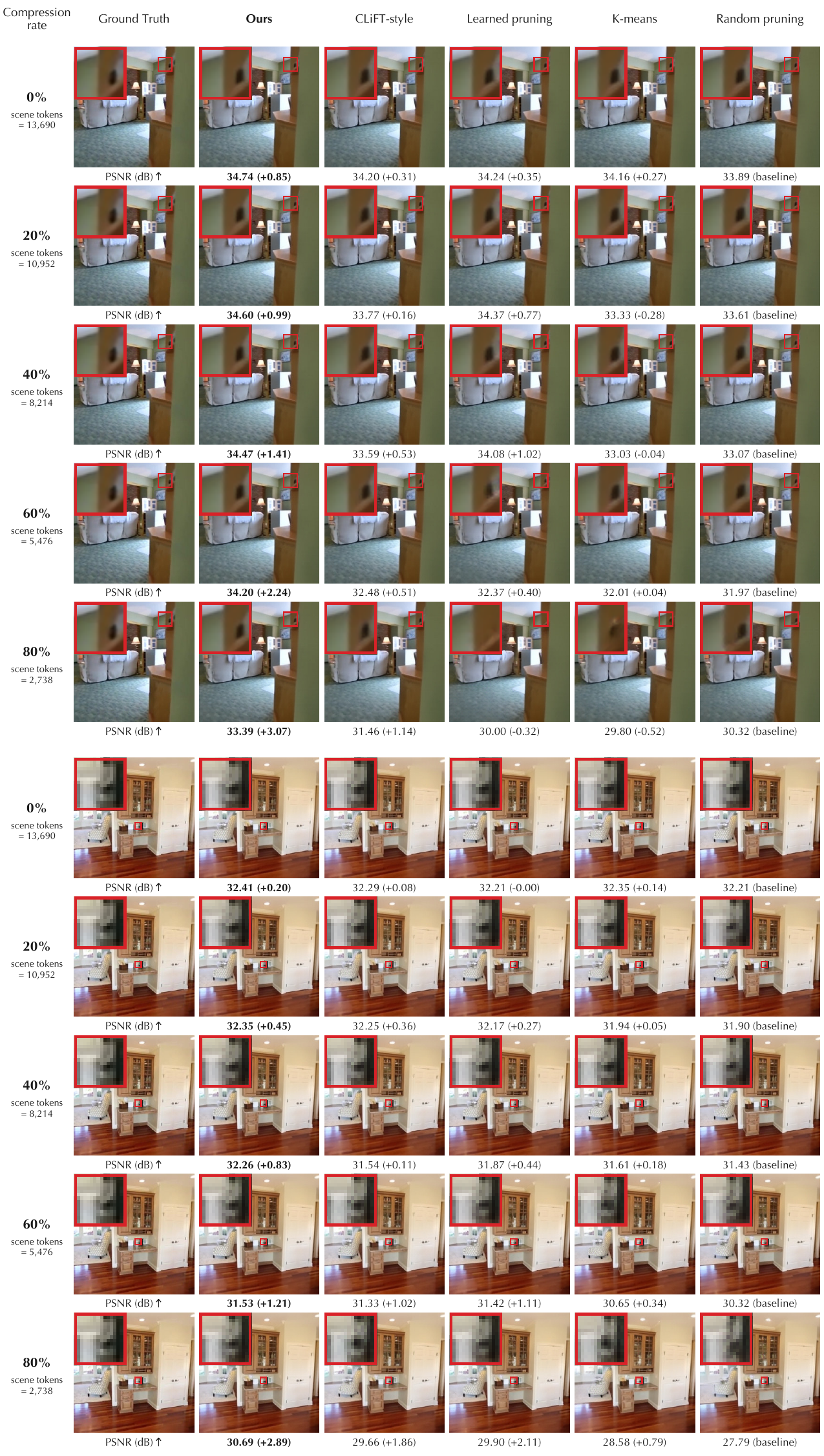}
\caption{Visual comparison of different condensers on RealEstate10K.}
\label{fig:compression-visual-05}
\end{figure}

\clearpage
\begin{figure}[p]
\centering
\includegraphics[width=\linewidth,height=0.95\textheight,keepaspectratio]{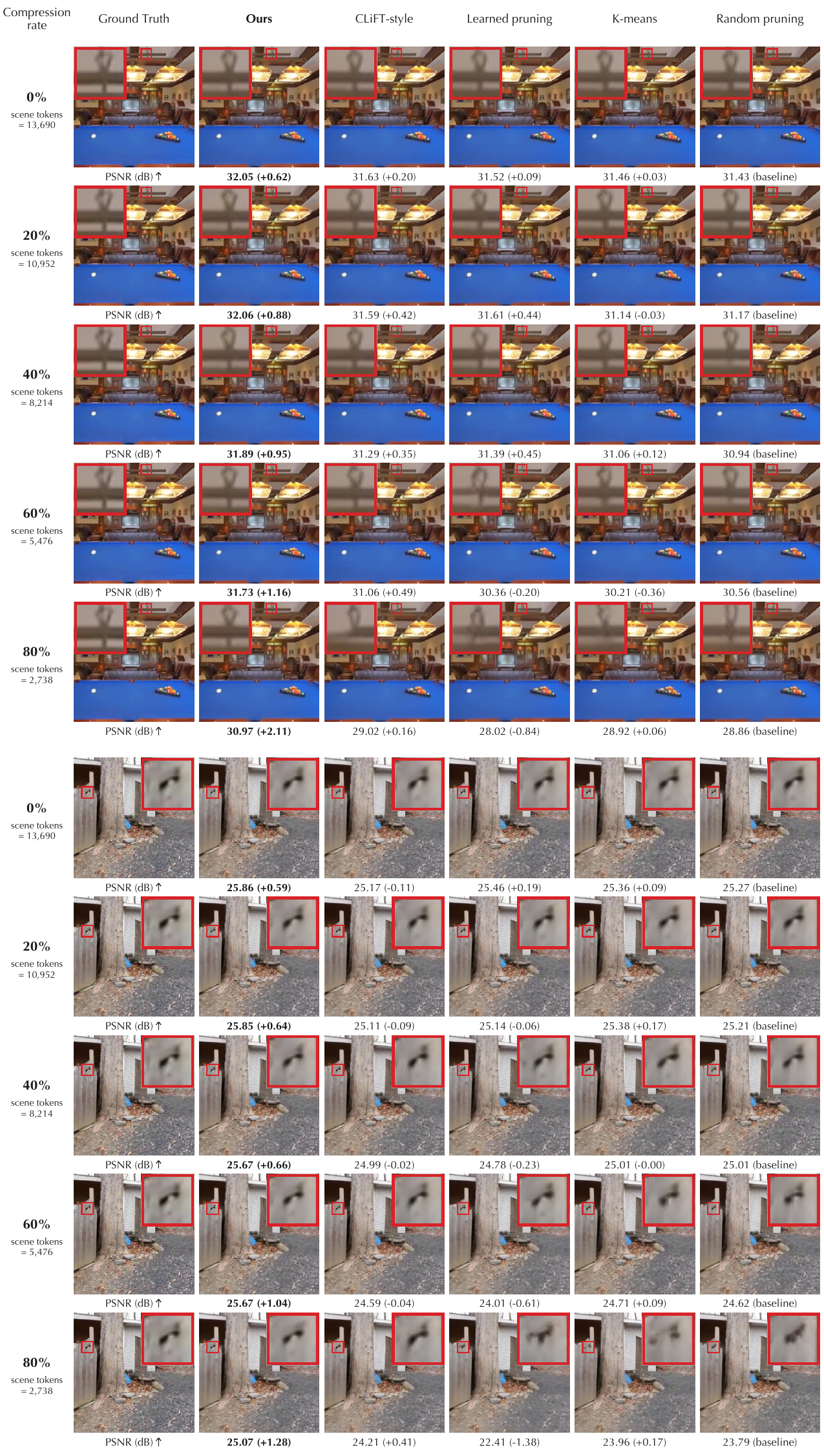}
\caption{Visual comparison of different condensers on RealEstate10K.}
\label{fig:compression-visual-06}
\end{figure}

\clearpage

%% file: sections/qualitative_selected.tex
\clearpage
\begin{figure}[p]
\centering
\includegraphics[width=\linewidth,height=0.95\textheight,keepaspectratio]{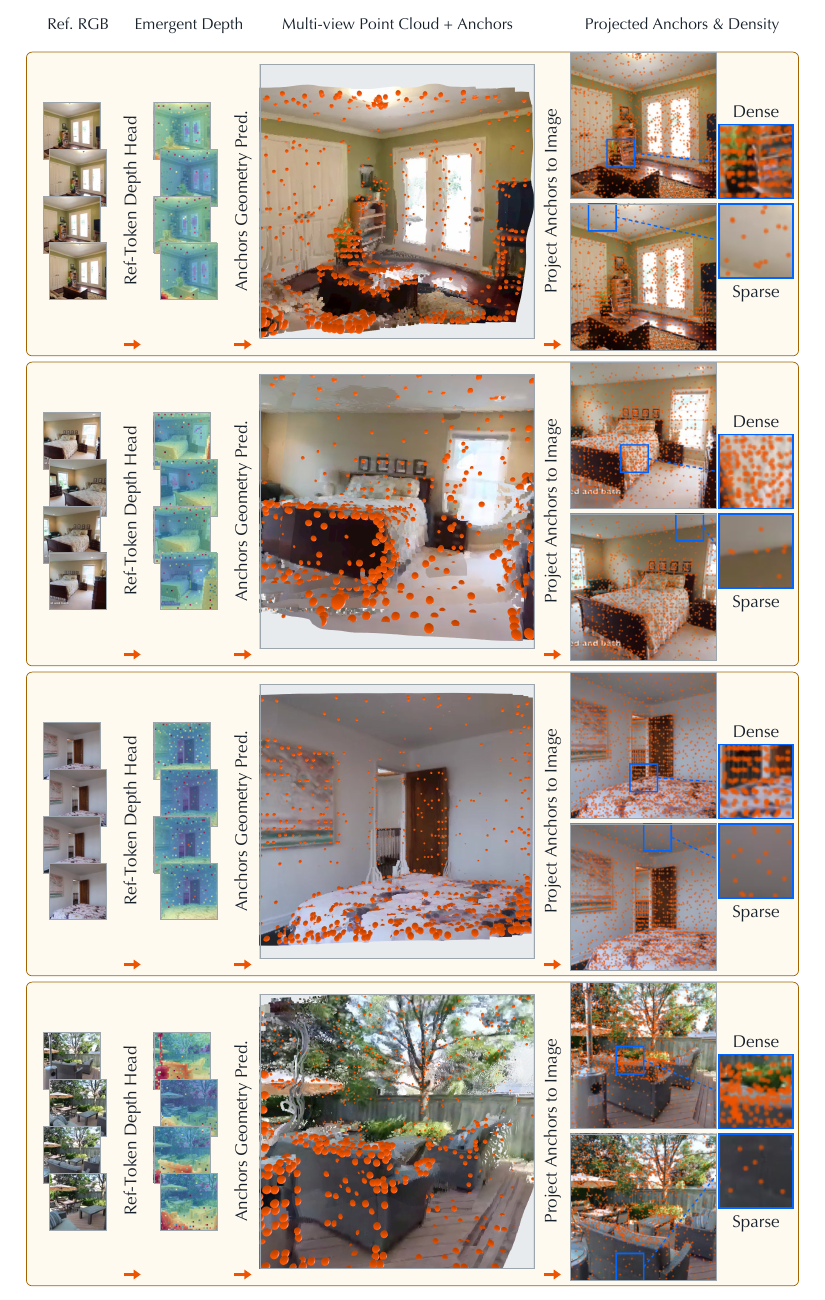}
\caption{Emergent geometry and geometry- and scene-complexity-aware anchor placement on RealEstate10K.}
\label{fig:qualitative-re10k-1}
\end{figure}

\clearpage
\begin{figure}[p]
\centering
\includegraphics[width=\linewidth,height=0.95\textheight,keepaspectratio]{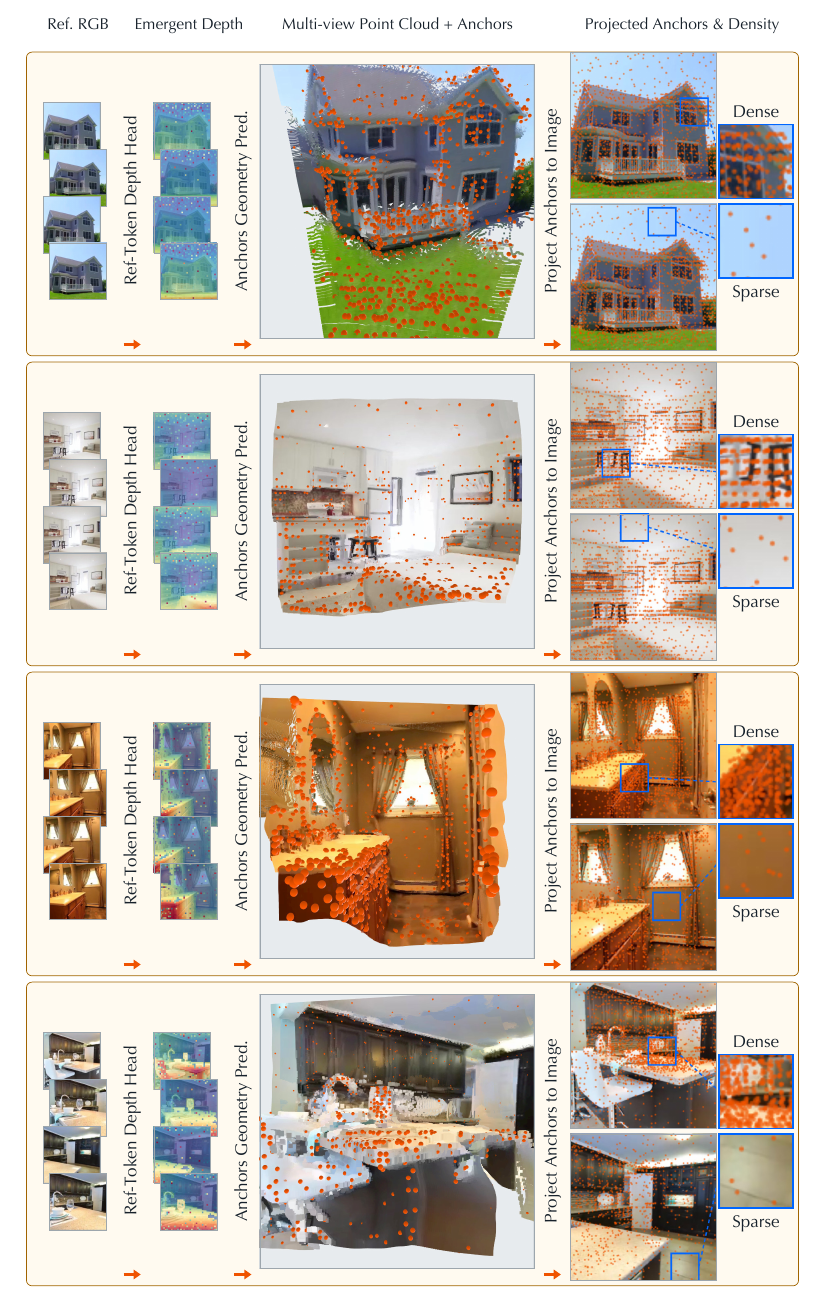}
\caption{Emergent geometry and geometry- and scene-complexity-aware anchor placement on RealEstate10K.}
\label{fig:qualitative-re10k-2}
\end{figure}

\clearpage
\begin{figure}[p]
\centering
\includegraphics[width=\linewidth,height=0.95\textheight,keepaspectratio]{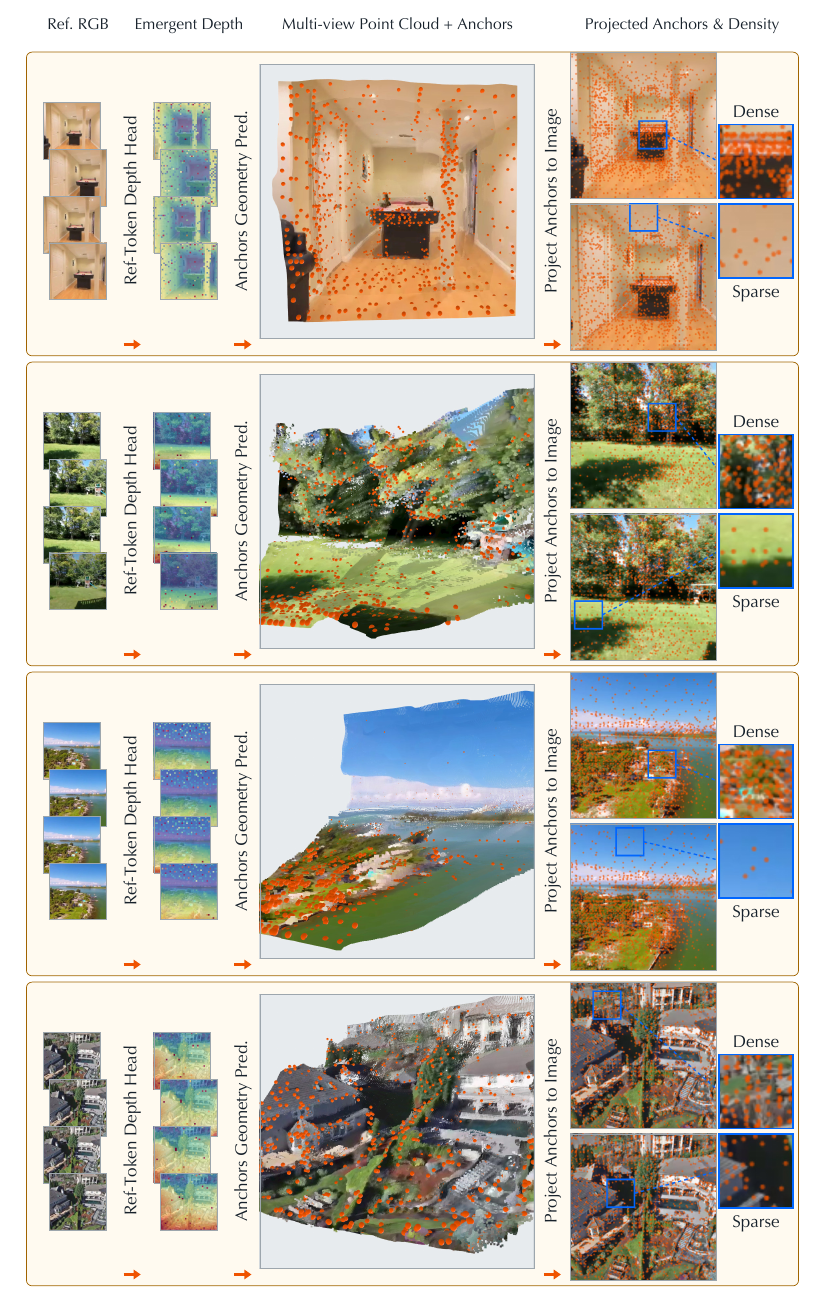}
\caption{Emergent geometry and geometry- and scene-complexity-aware anchor placement on RealEstate10K.}
\label{fig:qualitative-re10k-3}
\end{figure}

\clearpage
\begin{figure}[p]
\centering
\includegraphics[width=\linewidth,height=0.95\textheight,keepaspectratio]{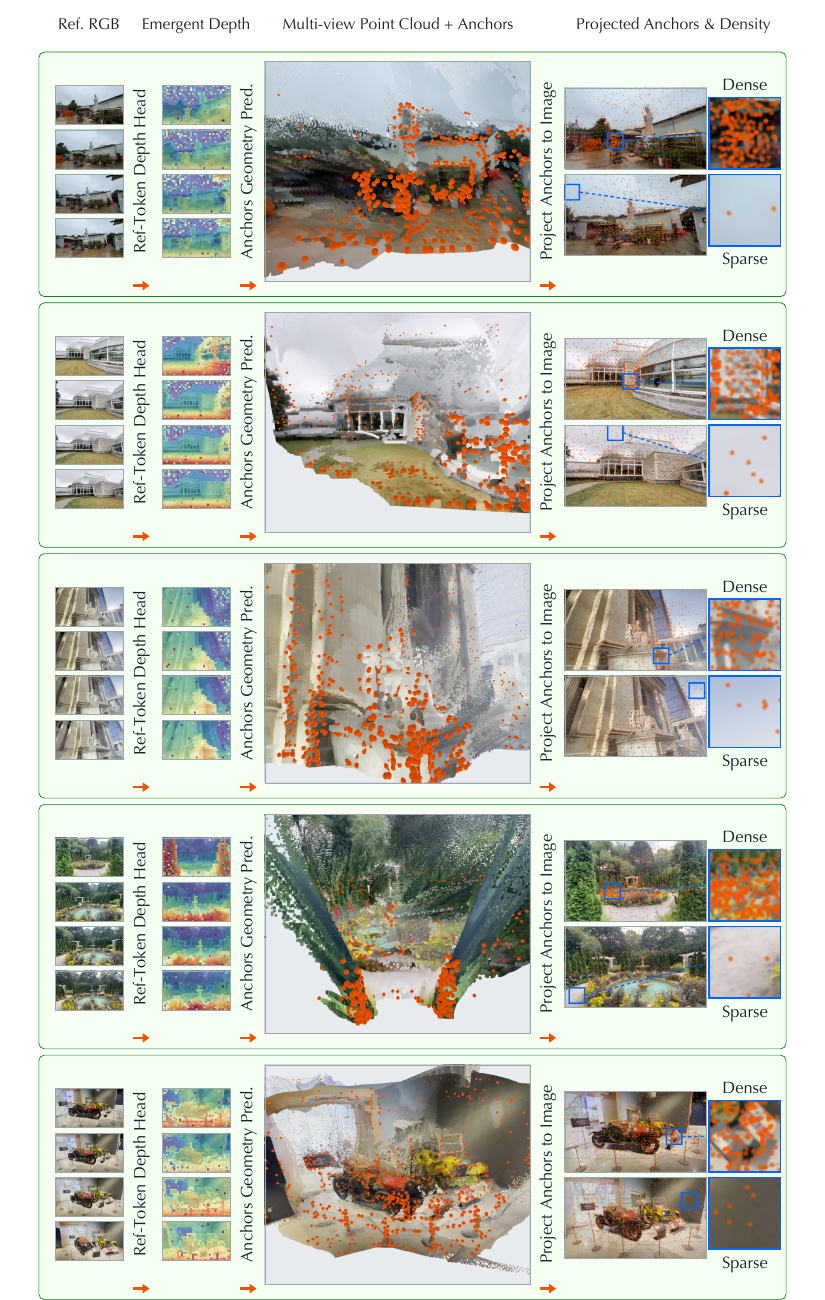}
\caption{Emergent geometry and geometry- and scene-complexity-aware anchor placement on DL3DV.}
\label{fig:qualitative-dl3dv-1}
\end{figure}

\clearpage
\begin{figure}[p]
\centering
\includegraphics[width=\linewidth,height=0.95\textheight,keepaspectratio]{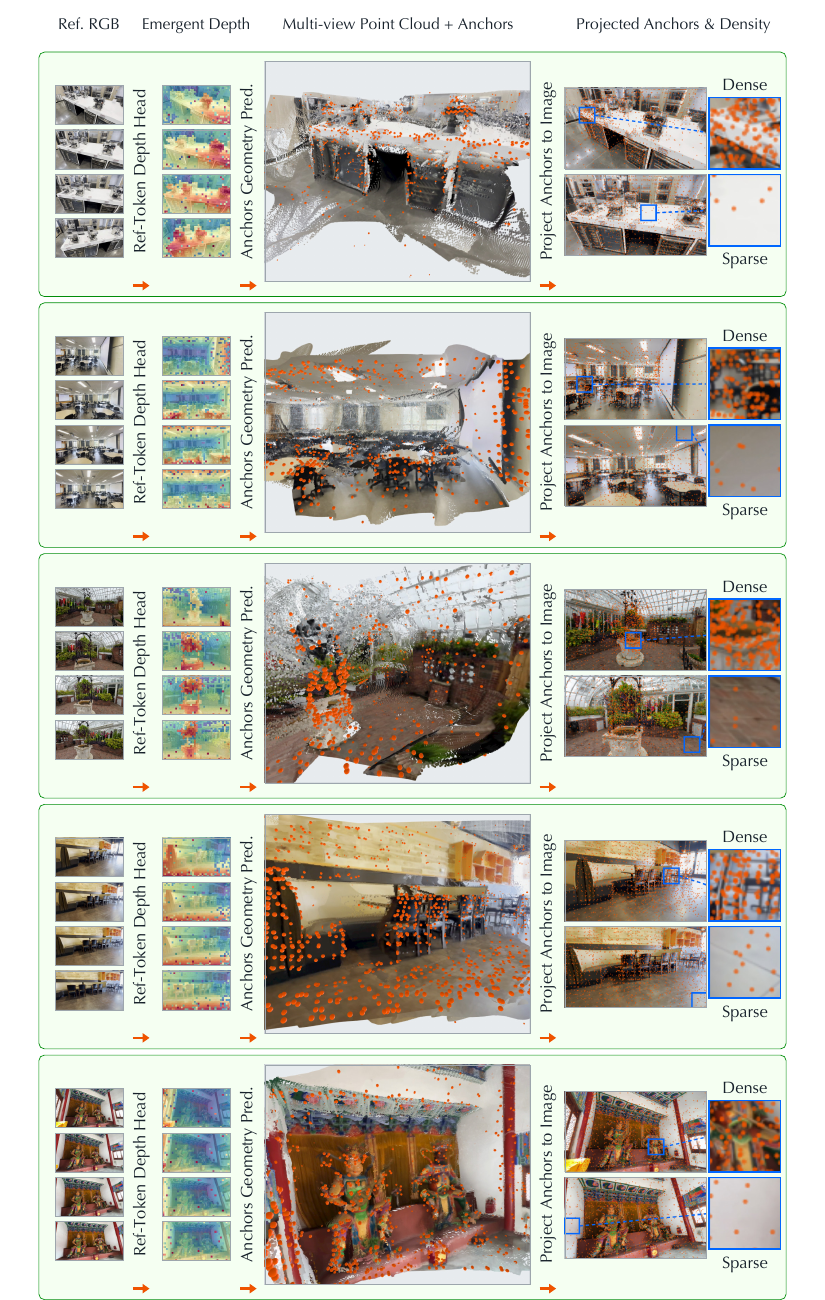}
\caption{Emergent geometry and geometry- and scene-complexity-aware anchor placement on DL3DV.}
\label{fig:qualitative-dl3dv-2}
\end{figure}

\clearpage
\begin{figure}[p]
\centering
\includegraphics[width=\linewidth,height=0.95\textheight,keepaspectratio]{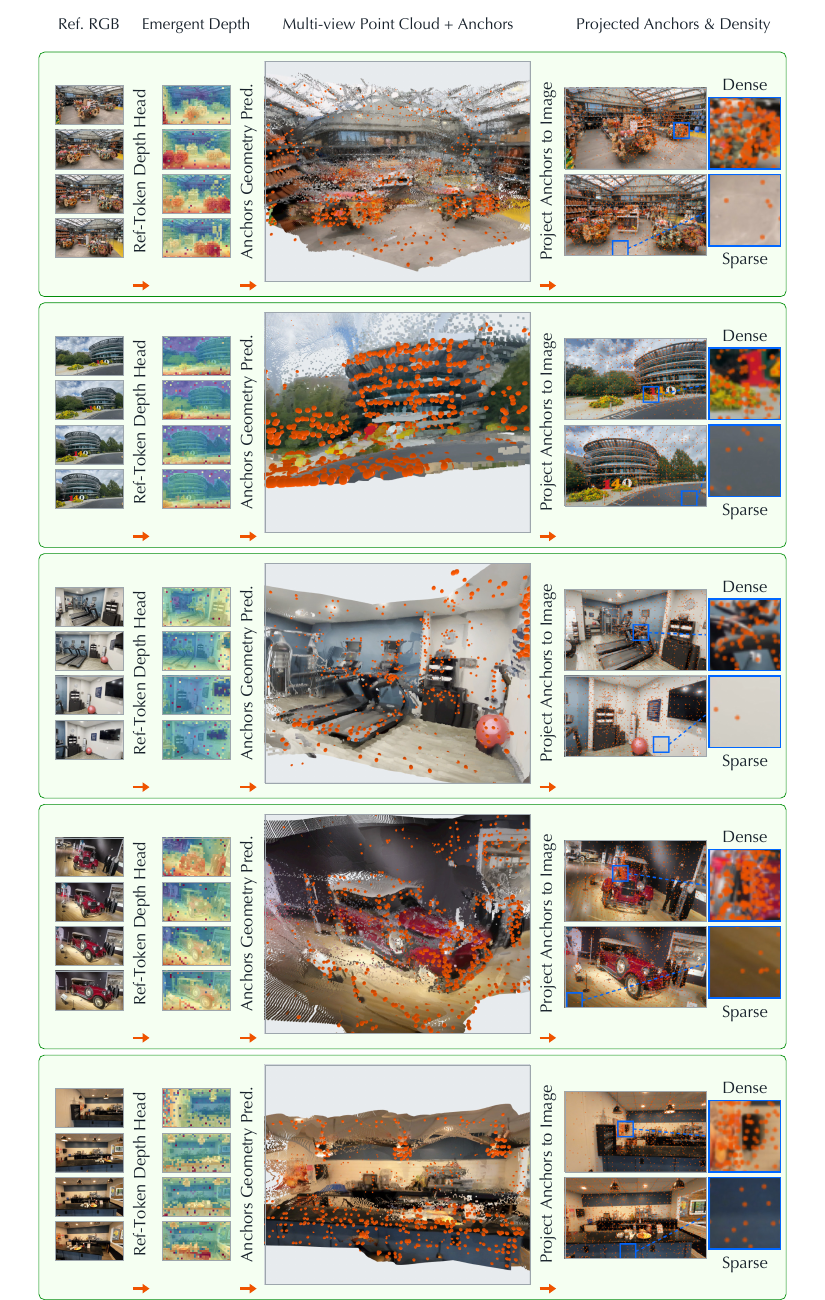}
\caption{Emergent geometry and geometry- and scene-complexity-aware anchor placement on DL3DV.}
\label{fig:qualitative-dl3dv-3}
\end{figure}

\clearpage